\documentclass[lettersize,journal]{IEEEtran}
\usepackage{amsmath,amsfonts}
\usepackage{algorithmic}
\usepackage{array}
\usepackage[caption=false,font=footnotesize,labelfont=rm,textfont=rm]{subfig}
\usepackage{textcomp}
\usepackage{stfloats}
\usepackage{url}
\usepackage{verbatim}
\usepackage{graphicx}
\usepackage{capt-of}
\usepackage{subfig}
\usepackage{hyperref}
\usepackage{multirow}
\usepackage{makecell}
\usepackage{xcolor}
\usepackage{cite}
\usepackage{ulem}
\usepackage{tabu}                      
\usepackage{booktabs}                  
\usepackage{lipsum}                    
\usepackage{mwe}                       
\usepackage{colortbl}
\usepackage{fancyhdr}
\def\BibTeX{{\rm B\kern-.05em{\sc i\kern-.025em b}\kern-.08em
    T\kern-.1667em\lower.7ex\hbox{E}\kern-.125emX}}
\usepackage{balance}

\fancypagestyle{firstpage}{
  \fancyhf{} 
  \fancyfoot[C]{\footnotesize © 2024 IEEE. Personal use is permitted, but republication/redistribution requires IEEE permission. \\See https://www.ieee.org/publications/rights/index.html for more information.} 

}

\begin{document}
\title{
Depth Completion with Multiple Balanced Bases and Confidence for Dense Monocular SLAM
}

\author{
Weijian Xie\IEEEauthorrefmark{1}, \and
Guanyi Chu\IEEEauthorrefmark{1}, \and Quanhao Qian\IEEEauthorrefmark{1}, \and Yihao Yu, \and Hai Li, \and Danpeng Chen, \and Shangjin Zhai, \\
\and Nan Wang, \and Hujun Bao, \and Guofeng Zhang\IEEEauthorrefmark{2}
\thanks{H. Bao and G. Zhang, H. Li are with the State Key Lab of CAD\&CG, Zhejiang University. E-mails: \{baohujun, zhangguofeng, garyli\}@zju.edu.cn.}
\thanks{W. Xie and D. Chen are with the State Key Lab of CAD\&CG, Zhejiang University. W. Xie is also with SenseTime Research, and D. Chen is also affiliated with Tetras.AI. E-mails: xieweijian@sensetime.com, chendanpeng@tetras.ai.}
\thanks{G. Chu, Q. Qian, Y. Yu, S. Zhai and N. Wang are with SenseTime Research. E-mails: \{chuguanyi1, qianquanhao1, yuyihao, zhaishangjin, wangnan\}@sensetime.com.}
\thanks{\IEEEauthorrefmark{1} Equal contribution}
\thanks{\IEEEauthorrefmark{2} Corresponding author}
\thanks{Digital Object Identifier 10.1109/TVCG.2024.3431926}
}


\maketitle
\thispagestyle{firstpage}
\newcommand{\xwjarxiv}[1]{\textcolor{black}{#1}}
\newcommand{\xwjtvcg}[1]{\textcolor{black}{#1}}
\newcommand{\xwjrevise}[1]{\textcolor{black}{#1}}
\newcommand{\qqhrevise}[1]{\textcolor{black}{#1}}
\newcommand{\xwjminor}[1]{\textcolor{black}{#1}}

\newcommand{\majorrevise}[0]{\color{black}}
\newcommand{\minorrevise}[0]{\color{black}}

\begin{abstract}

Dense SLAM based on monocular cameras does indeed have immense application value in the field of AR/VR, especially when it is performed on a mobile device. In this paper, we propose a novel method that integrates a light-weight depth completion network into a sparse SLAM system using a multi-basis depth representation, so that dense mapping can be performed online even on a mobile phone.
Specifically, we present a specifically optimized multi-basis depth completion network, called BBC-Net, tailored to the characteristics of traditional sparse SLAM systems. BBC-Net can predict multiple balanced bases and a confidence map from a monocular image with sparse points generated by off-the-shelf keypoint-based SLAM systems. 
The final depth is a linear combination of predicted depth bases that can be \xwjminor{easily} optimized by tuning the corresponding weights. To seamlessly incorporate the weights into traditional SLAM optimization and ensure efficiency and robustness, we design a set of depth weight factors, which makes our network a versatile plug-in module, facilitating easy integration into various existing sparse SLAM systems and significantly enhancing global depth consistency through bundle adjustment.
To verify the portability of our method, we integrate BBC-Net into two representative SLAM systems. The experimental results on various datasets show that the proposed method achieves better performance in monocular dense mapping than the state-of-the-art methods. We provide an online demo running on a mobile phone, which verifies the efficiency and mapping quality of the proposed method in real-world scenarios.
\end{abstract}

\begin{IEEEkeywords}
Depth Completion, Dense SLAM, Multi-Basis
\end{IEEEkeywords}

  

\section{Introduction}
 \label{sec:intro}

\IEEEPARstart{R}{eal-time} 6-DoF pose estimation and reconstruction of the surrounding environment is a fundamental challenge for AR/VR. 
\xwjarxiv{Current monocular visual~(-inertial) odometry or SLAM methods~\cite{campos2021orb,forster2014svo,qin2018vins,liu2016robust,zubizarreta2020direct} based on multi-view geometry have achieved considerable success in pose estimation. However,} these methods can only generate sparse point clouds or semi-dense maps, \xwjrevise{posing challenges in handling complex AR applications, such as collision detection and occlusion handling.}
\xwjrevise{D}ense SLAM methods~\cite{izadi2011kinectfusion, whelan2015elasticfusion, bylow2013real, schops2019bad}, \xwjrevise{capable of producing real-time dense meshes,} often rely on additional sensors like Time-of-Flight (ToF), LiDAR, etc. Considering the economic cost, an ideal approach is to generate dense meshes using a monocular camera, a highly challenging problem, especially on resource-constrained mobile devices.

\begin{figure}[tb]
 \centering 
 \includegraphics[width=\columnwidth]{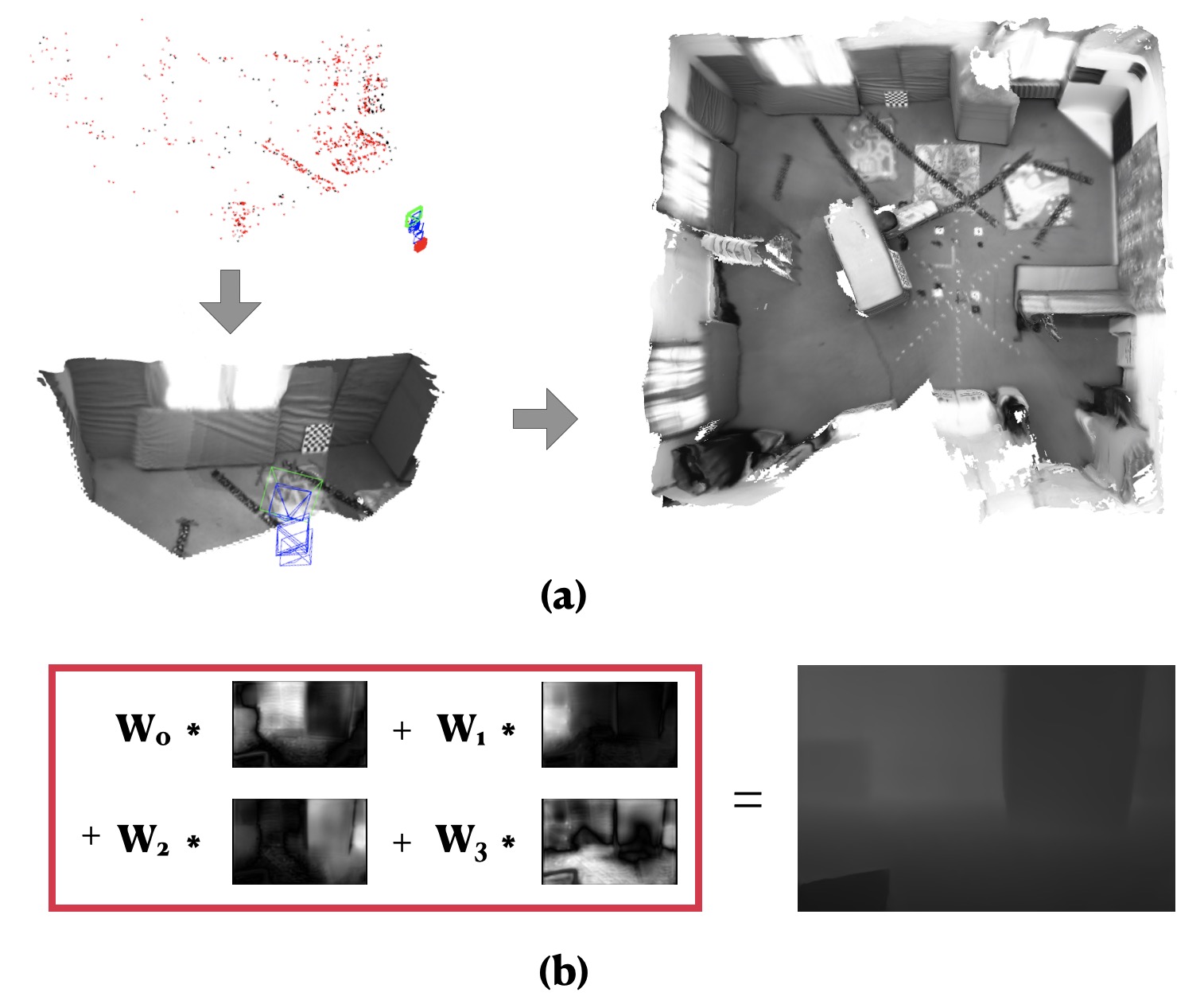}
 \caption{\xwjminor{(a) The real-time dense mapping process of BBC-ORBSLAM. (b) The final depth is the weighted linear sum of predicted depth bases. BBC-Net complements the sparse depth generated by traditional sparse SLAM(top left of (a)) to produce dense mesh(bottom left of (a)). By incorporating the weights of depth bases into SLAM optimization, a globally consistent mesh is obtained (right of (a)).}}
 \label{fig:story_image}
\end{figure}
\xwjarxiv{With the development of deep learning, several end-to-end methods~\cite{murez2020atlas,tang2018ba, sun2021neuralrecon, sucar2021imap, zhu2022nice, teed2021droid, rosinol2023probabilistic} have emerged to tackle dense or semi-dense mapping from a monocular camera. However, these methods often demand substantial computational resources, making them unsuitable for mobile devices. Traditional sparse SLAM methods remain the mainstream solution for real-time pose estimation due to their efficiency in practical applications.}
\xwjminor{Depth prediction~\cite{bhat2023zoedepth} from single image typically suffer from low accuracy and difficulty in ensuring global consistency. A more natural approach is to integrate a depth completion ~\cite{ma2018sparse, warburg2022self, qu2020depth, jaritz2018sparse, yan2022rignet, warburg2022sparseformer, cheng2019learning, wong2021unsupervised, cheng2020cspn++,zhao2021adaptive,hu2021penet,park2020non,zhang2023completionformer}} with a traditional monocular SLAM framework.
However, previous depth completion networks, such as \cite{qu2020depth}, are primarily designed for completing semi-dense depth captured by depth sensors, typically handling noise-free or minimally noisy randomly sparse depths. In SLAM systems, sparse depth often originates from sparser fast corner points with increased noise. Consequently, the design of a depth completion network for integration with a SLAM system necessitates essential customization of the network.

\begin{figure}[tb]
 \centering 
 \includegraphics[width=\columnwidth]{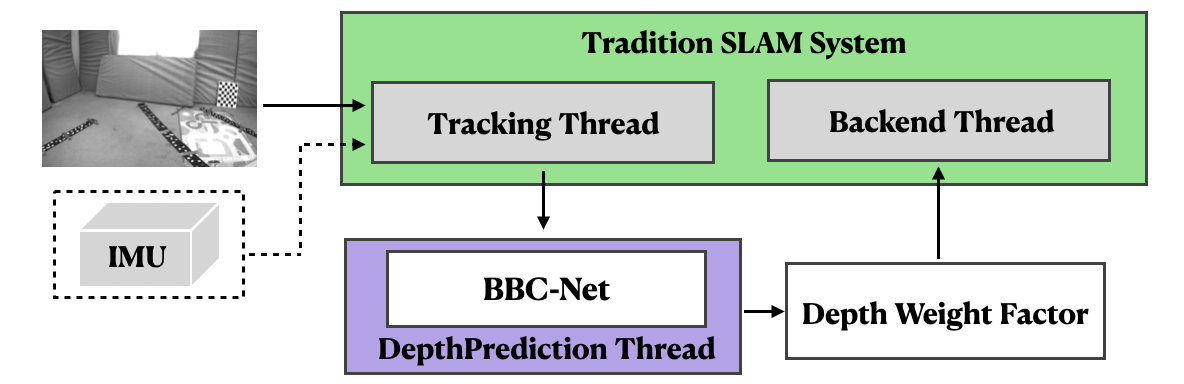}
 \caption{\xwjarxiv{General framework. \xwjrevise{The depth completion network uses images and sparse points from the tracking module as input, predicting depth bases and confidences. The weights of the depth bases are then integrated into the SLAM optimizer for continuous depth optimization.}}}
 \label{fig:genera_system_overview}
\end{figure}

Some methods~\cite{tateno2017cnn,yang2020d3vo,sun2022improving,yan2021monocular} have aimed to utilize the predicted depth to enhance SLAM accuracy. However, these methods cease to optimize the depth once generated, or just align the scale of the predicted depth, which makes the global consistency of the depth cannot be maintained. GeoRefine~\cite{ji2022georefine} incorporates RAFT-flow~\cite{teed2020raft} to enhance the front-end tracking and utilizes self-supervised learning to improve depth accuracy, making it evidently a non-lightweight solution.
Methods in~\cite{bloesch2018codeslam,zuo2021codevio,matsuki2021codemapping} use a variational autoencoder to compress dense depth into low-dimensional codes, \xwjrevise{integrable into traditional SLAM optimization for dense depth refinement}. 
\xwjrevise{A significant drawback is their heavy dependence on GPU-based computations, necessitating GPU usage not only during Jacobian matrix calculation but also after each optimization iteration for reinferring new dense depths.} 
\xwjarxiv{In this work, we propose a highly flexible framework (as shown in \autoref{fig:genera_system_overview}) to seamlessly transform a traditional sparse SLAM into a dense SLAM. With minimal modifications to the existing sparse SLAM, our method can efficiently achieve this conversion without introducing significant additional computational burden.}
\xwjarxiv{Similar to CodeVIO~\cite{zuo2021codevio}, CodeMapping~\cite{matsuki2021codemapping}, we incorporate the network-predicted depth into the optimization of traditional SLAM for achieving global consistent depths. The distinction lies in our use of a multi-basis depth representation, avoiding redundant network inferences and rendering our method highly efficient, particularly for mobile applications.}
\xwjrevise{While ~\cite{qu2020depth} introduced multi-basis depth representation to improve the quality of single-frame depth completion, we leverage this representation for its distinct advantage in seamlessly integrating with traditional SLAM optimization, presenting a different motivation. Moreover, the integration of depth completion results into a SLAM system introduces novel and more intricate challenges(see \autoref{sec:net}), aspects that were not considered by \cite{qu2020depth}.}
\xwjarxiv{To the best of our knowledge, we are the first to integrate multi-basis depth representation with traditional SLAM systems.}

The main contributions of this paper are listed as follows:

\begin{itemize}
\item 
We propose a light-weight multi-basis depth completion network, named BBC-Net, which is specifically designed for achieving dense monocular SLAM, by addressing the issue of imbalance in bases and predicting balanced bases with confidence simultaneously.
\item
\xwjrevise{We propose a set of factors based on the multi-basis depth representation, enabling the optimization of the predicted depth. The flexibility to combine these factors to ensure precision and robustness in optimization, and we can even use marginalization to expedite the optimization.}
\item 
We propose a complete framework to integrate the proposed BBC-Net as a plugin into traditional sparse monocular SLAM systems, thus transforming them into dense SLAM systems. 
To illustrate the transformation from sparse SLAM to dense SLAM, we have chosen two representative sparse SLAM systems, ORB-SLAM3~\cite{campos2021orb} and VINS system (introduced in Sec.~\ref{sec:BBC-VINS}), as exemplar cases to showcase our integration framework.
\item
We develop an online demo that runs on a mobile phone, which verifies the efficiency of our method and demonstrates the mapping quality in real-world scenarios.
\end{itemize}

The experiments on various datasets demonstrate that our method achieves better performance than the state-of-the-art in monocular dense mapping. 
\autoref{fig:story_image} shows an example.
\textcolor{black}{The source code of BBC-Net can be found at \href{https://github.com/zju3dv/BBCNet}{https://github.com/zju3dv/BBCNet}.}

\section{Related Work}
\label{sec:reference}

\subsection{Depth Completion}
\xwjrevise{
Depth completion networks are primarily designed to fill in semi-dense depth captured by depth sensors. Therefore, previous depth completion methods typically dealt with sparse depth obtained from the ground truth depth or semi-dense depth collected by sensors through random sampling.
Ma et al.~\cite{ma2018sparse} proposed a ResNet-based model for early fusion, combining RGB images and sparse depth points as a single input. 
Late fusion methods~\cite{jaritz2018sparse, yan2022rignet, warburg2022sparseformer, wong2021unsupervised} use different encoder networks to extract features from RGB images and depth, concatenating these features to predict depth.
Qu et al.~\cite{qu2020depth} represented the final depth as a weighted sum of implicit depth bases, enhancing accuracy by replacing the final 1×1 CNN layer with a least squares algorithm. A more widely adopted approach~\cite{cheng2019learning, cheng2020cspn++, park2020non} to improve accuracy involves incorporating an additional SPN network at the end of the UNet architecture to further refine the final depth.
Despite the relatively high accuracy of sparse depth from depth sensors, depth completion networks tend to overfit the input sparse depths. However, sparse points from SLAM systems differ significantly from randomly sampled sparse depth from ground truth. Sparse points from SLAM are often fast corners, typically located at boundaries where the depth of surrounding pixels is prone to abrupt changes, and they exhibit greater noise. In such scenarios, overfitting to the input sparse depth may have negative consequences.
}

\subsection{VIO/SLAM}
PTAM~\cite{klein2007parallel} is the first keyframe-based multi-thread SLAM system. Keyframe-based methods can do optimization with only a few frames, making it a cost-effective method. After that, almost all the following SLAM framework runs a real-time local tracking thread, which we call frontend, and a slower global consistency mapping thread, which call backend. 
According to the way of front-end tracking, SLAM can be divided into direct~\cite{engel2017direct, engel2014lsd}, indirect~\cite{mur2015orb, liu2016robust}, and semi-direct~\cite{forster2014svo, forster2016svo} systems. 
In the SLAM framework, the backend maintains a global map of points and keyframes. ORB-SLAM is one of the most outstanding indirect method SLAM frameworks in the last decades. The back-end of ORB-SLAM preserves the relationship between all corresponding keyframes and continuously optimizes both map points and keyframes by bundle adjustment. In recent years, with the promotion of low-cost IMU, VINS has been widely used. A VINS (visual-inertial system)~\cite{qin2018vins, mourikis2007multi, Geneva2020ICRA} usually consists of a small sliding window as frontend and combined with a lightweight pose-graph-based backend, which will not save the correspondence of keyframes and only optimize the keyframe poses through loop closure. 
In this work, to better verify our methods, we implement our method both in a full backend of ORB-SLAM3 and a lightweight backend of VINS.

\subsection{Deep Dense Mapping}

~\cite{murez2020atlas,tang2018ba} present end-to-end approaches for addressing the challenge of monocular dense mapping.~\cite{sun2021neuralrecon} focus on reconstructing dense meshes from provided images and poses. \xwjrevise{~\cite{sucar2021imap, zhu2022nice} present end-to-end implicit dense SLAM with RGBD input.} ~\cite{rosinol2023probabilistic} builds upon the work of~\cite{teed2021droid} by introducing dense reconstruction. However, all these methods require large amounts of computation resources.

In recent years, with the development of depth prediction, many researches have focused on combining depth prediction results with SLAM systems.
Sun et al.~\cite{sun2022improving} designed two depth prediction networks, one for predicting relative depths to improve Visual Odometry accuracy, and another for predicting scale-consistent depths for dense mapping. Liu et al.~\cite{liu2022depth} utilize the predicted depth to improve the initialization and map point triangulation. Yang et al.~\cite{yang2020d3vo} incorporate the outputs of depth prediction networks and pose prediction networks into the DSO~\cite{engel2017direct} to address the issue of scale ambiguity. However, the problem with these methods is that once the depth is predicted by the network will not be optimized anymore. 
CNN-SLAM~\cite{tateno2017cnn} predicts depth from a single frame and uses dense depth alignment to generate the 3D dense map. 
Koestler et al.~\cite{koestler2022tandem} propose a CVA-MVSNet based on MVSNet~\cite{yao2018mvsnet} to predict depth from several pose-given images, and the depth will be fused into TSDF. The front-end tracking depends on the updated TSDF, which leads to the low efficiency of the method. Another creative way to continuously optimize the depth in the BA process is variational auto-encoders~\cite{bloesch2018codeslam,zuo2021codevio,matsuki2021codemapping,czarnowski2020deepfactors}. However, these methods are heavy reliance on GPU, limiting the implementation on mobile. 


\section{BBC-Net}
\label{sec:net}

\begin{figure}[ht]
\centering
\subfloat [\xwjarxiv{Multi-basis predicted by BBC-Net, scaled with the same factor.}] {
\includegraphics[width=0.23\columnwidth]{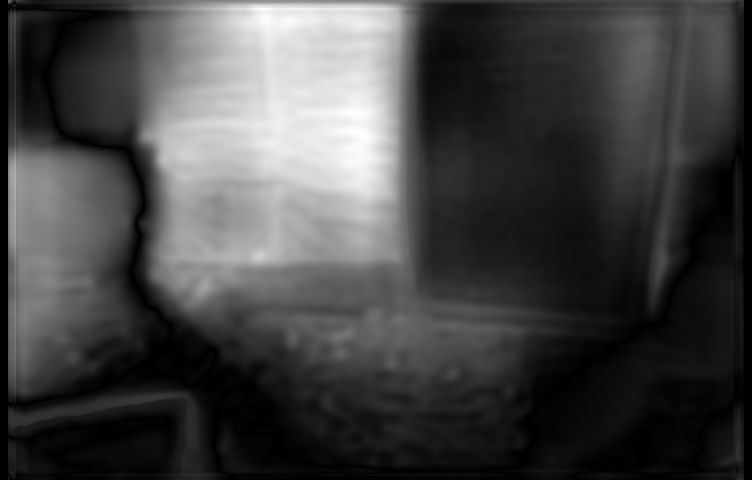}
\includegraphics[width=0.23\columnwidth]{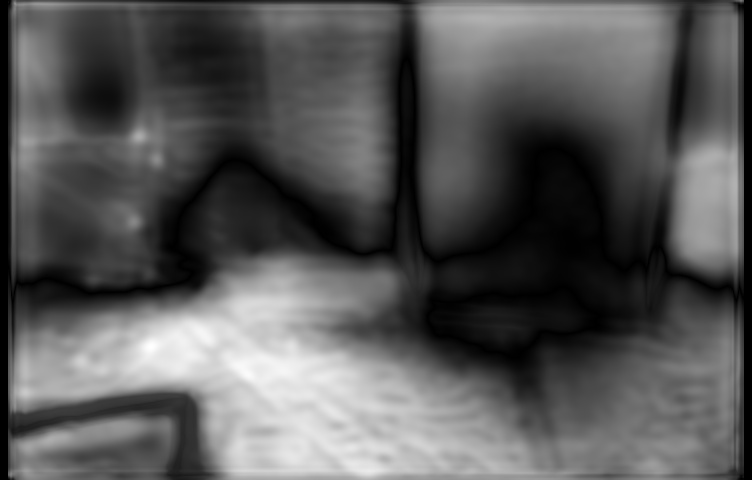}
\includegraphics[width=0.23\columnwidth]{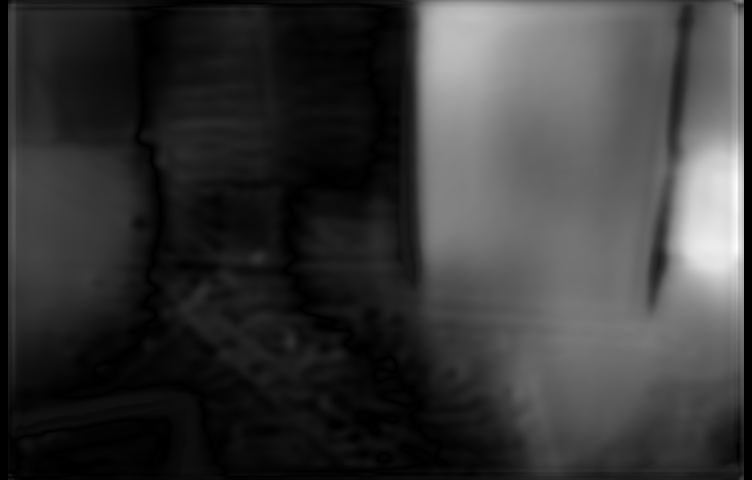}
\includegraphics[width=0.23\columnwidth]{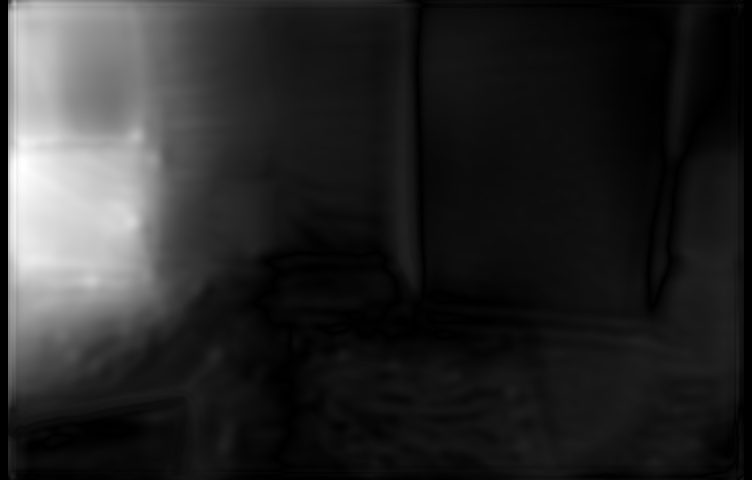}
} \\
\subfloat [\xwjarxiv{Multi-basis predicted by~\cite{qu2020depth}, scaled with the same factor.}] {
\includegraphics[width=0.23\columnwidth]{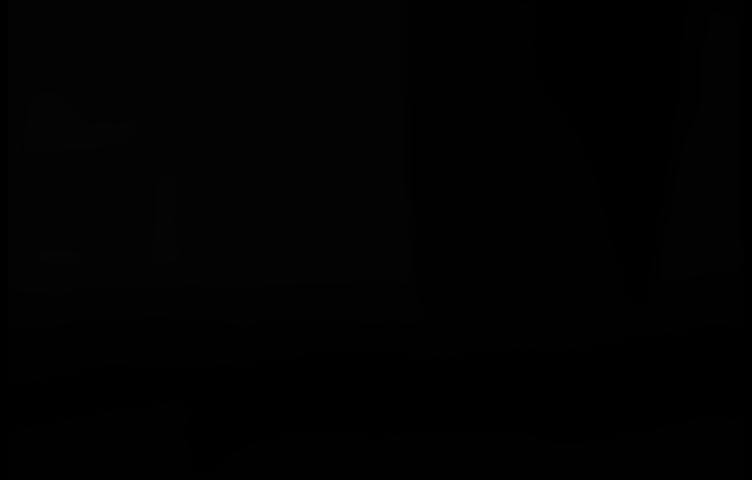}
\includegraphics[width=0.23\columnwidth]{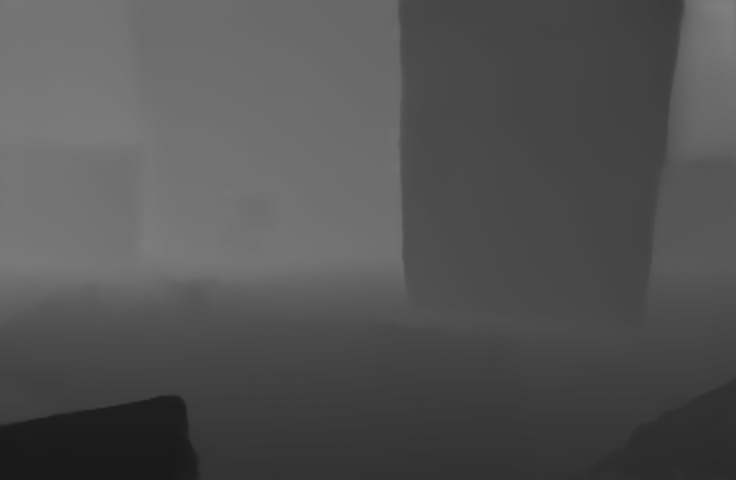}
\includegraphics[width=0.23\columnwidth]{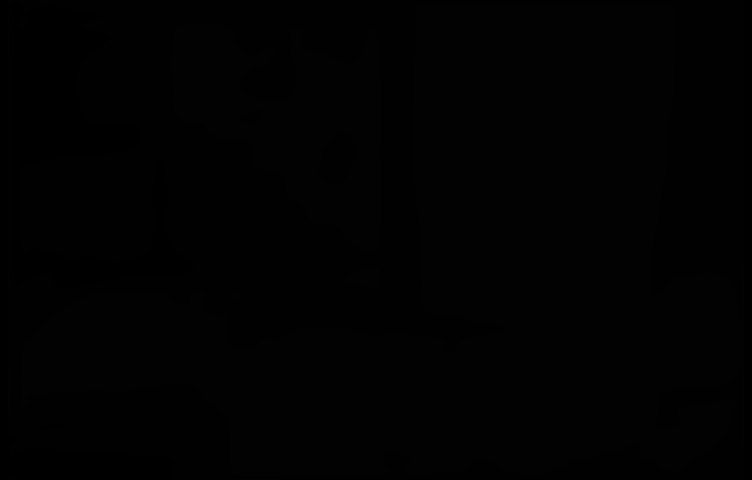}
\includegraphics[width=0.23\columnwidth]{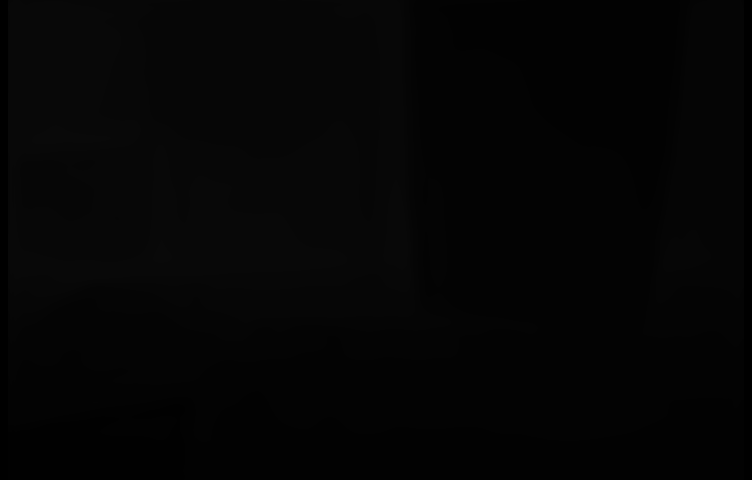}
} \\
\caption{\xwjarxiv{(a) The depth bases predicted by our network exhibit excellent information distribution, with each base learning the relative depth of a region.
(b) The depth bases generated by \cite{qu2020depth} exhibit obvious imbalance, with typically only one base output containing almost all the information.
}
}  
\label{fig:bases}
\end{figure}

\begin{figure}[!htb]
\centering
\subfloat [\xwjarxiv{Input image}] {
\includegraphics[width=0.3\columnwidth]{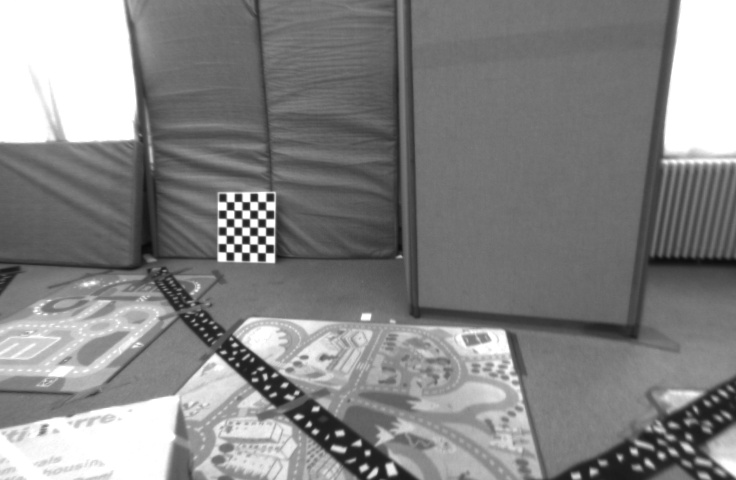}
}
\subfloat [\xwjarxiv{Depth (~\cite{qu2020depth})}] {
\includegraphics[width=0.3\columnwidth]{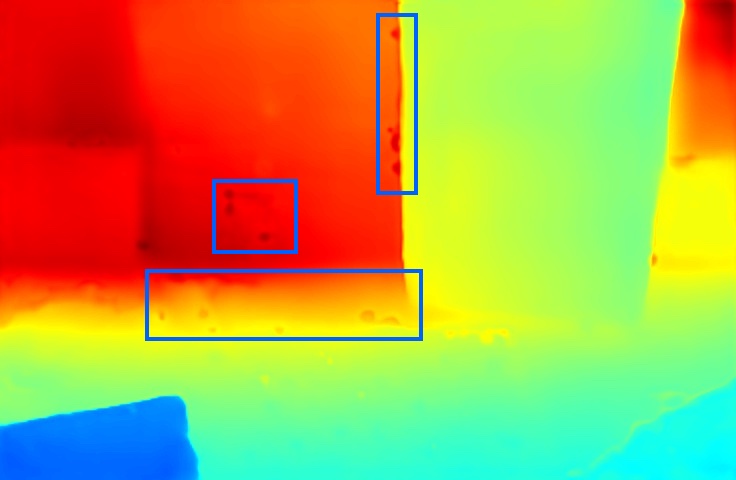}
}
\subfloat [\xwjarxiv{Depth (BBC-Net)}] {
\includegraphics[width=0.3\columnwidth]{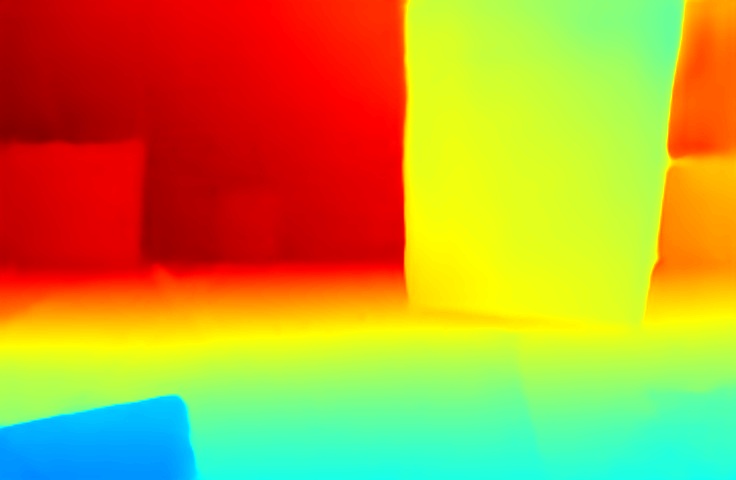}
} \\
\subfloat[\xwjarxiv{Mesh generated by (b)}]{
\includegraphics[width=0.46\columnwidth]{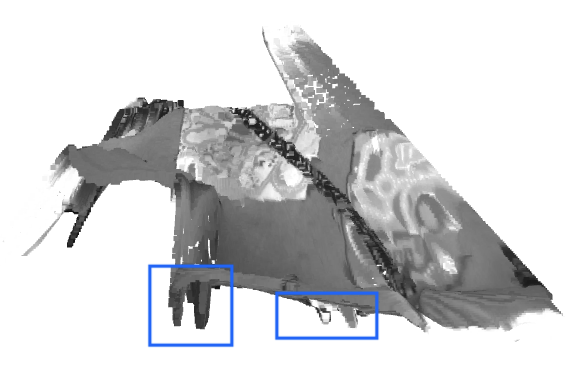}
}
\subfloat[\xwjarxiv{Mesh generated by (c)}]{
\includegraphics[width=0.46\columnwidth]{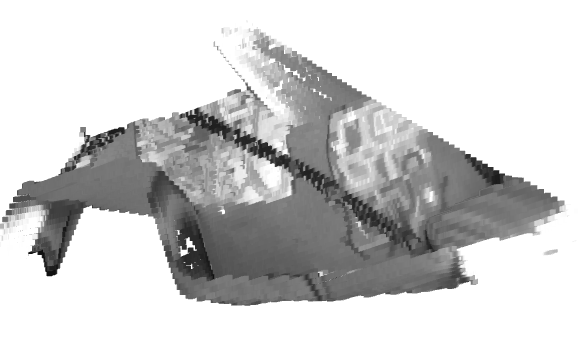}
}
\caption{ 
Outliers are highlighted with blue boxes.
Although outliers constitute a small proportion of the predicted depths, they are clearly visible as prominent protrusions in the corresponding 3D mesh.
}  
\label{fig:BBC-Net_improvement}
\end{figure}

\begin{figure*}[tb]
 \centering 
 \includegraphics[width=\textwidth]{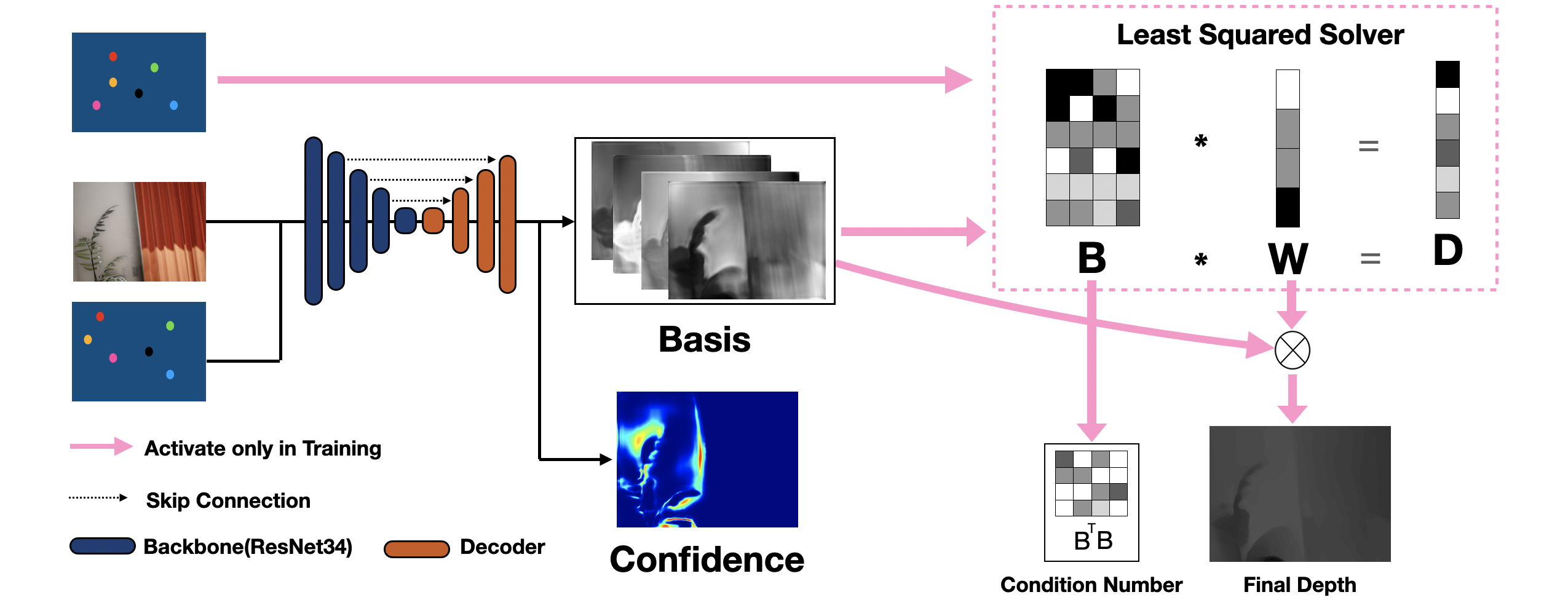}
 \caption{BBC-Net architecture. The network takes a grayscale image and sparse depth as inputs and outputs a set of depth bases and a confidence map. The pink arrow is activated only in the training process.}
 \label{fig:net}
\end{figure*}


\xwjrevise{Depth completion networks are prone to overfitting around the input sparse depth, which is not problematic in scenarios with small noise, sufficient quantity, and well-distributed sparse points. However, in SLAM systems, sparse points are typically distributed in areas prone to depth discontinuities, with high noise, extremely sparse quantity. Overfitting to the sparse depth inferred at this stage might adversely affect, even resulting in abnormal protrusions.
Similar situations are also discussed in \cite{matsuki2021codemapping}. 
In the specific case of the multi-basis depth representation method, this issue manifests as an imbalance of depth bases, exhibiting two aspects: 
}
firstly, there is a notable imbalance in the information learned by each depth base, with typically only one depth base outputting almost all of the information (as shown in \autoref{fig:bases}); secondly, there is also an imbalance in the learned information within each individual depth base.
\xwjrevise{
From the standpoint of nonlinear optimization, it is evident that under noisy conditions, base imbalance problem is more likely to result in optimization difficulties(see abnormal protrusions in \autoref{fig:BBC-Net_improvement}).
Moreover, integrating predicted depth with SLAM introduces another crucial consideration—dealing with the inevitable regions of large prediction errors. It is essential to prevent these regions from negatively impacting SLAM. Including depth points with significant errors during optimization may lead to decreased accuracy or even divergence. One straightforward solution is for the network to predict confidence alongside depth.
}

 To overcome the above problems, we propose BBC-Net, a multi-basis network designed to integrate with SLAM systems. As illustrated in \autoref{fig:net}, BBC-Net consists of two modules: an encode-decoder network and a linear least-squares solver module, similar to the approach presented in~\cite{qu2020depth}. The linear least square module estimates the weight of bases with the generated bases and the input sparse depths during the training process. 
 The network takes a grayscale image and sparse depths as inputs, yielding two branches of output: an N-dimensional vector of bases, matching the input resolution, and a corresponding confidence map for the input image. 
 \xwjrevise{
 As}~\cite{qu2020depth} has not released their code, we implemented their method as a baseline and further develop our network based on it. Here, we will \xwjrevise{focus on presenting} the main improvements we made compared to the baseline. \xwjrevise{For additional implementation details regarding both the baseline and BBC-Net, please refer to~\cite{qu2020depth} and our supplementary materials.} 

 Compared to~\cite{qu2020depth}, our approach makes several improvements. Firstly, we introduced a confidence loss along with an additional confidence output to prevent points with significant errors from influencing optimization and mapping. \xwjminor{We use confidence to remove outliers, and points with confidence below the threshold will not participate in optimization and dense mapping.} Secondly, we incorporated the bases balance loss, enhancing the supervised training process to mitigate overfitting to the input sparse depths.
 Our loss function is defined as follows:

\begin{equation}
\label{equa:loss}
    L = L_{d} + w_{c} L_{c} + w_{b} L_{b},
\end{equation}
 where $L_{d}$ is the depth consistency loss, $L_{c}$ is confidence loss, $L_{b}$ is bases balance loss, $w_{c}$ and $w_{b}$ is the weight of $w_c$ and $w_b$. In this paper, $w_c$ is set to 2, while $w_b$ is set to 0.1

The depth consistency loss can be simply defined as follows:
\begin{equation}
\label{equa:loss_depth}
  L_{d} = ||(D_{gt} - D_{pred}) ||,
\end{equation}
\xwjminor{where $D_{gt}$ is the ground truth depth, while $D_{pred}$ is predicted depth estimate by least squared solver.}

\textbf{Confidence Loss} 
 The design principle of the confidence loss is to assign higher confidence to areas with high accuracy in depth prediction, while penalizing high confidence in regions with significant depth prediction errors.
Inspired by~\cite{choi2021adaptive}, we define the confidence loss as follows:



\begin{equation}
\label{equa:loss_conf}
    L_{c} = ||(D_{gt} - D_{pred})\circ C||_1 + w_0\left \|  \frac{1}{C + 1}\right \| _1 ,
\end{equation}
where C is the confidence of the predicted depth. $\left \|  \frac{1}{C + 1}\right \|_1$ is designed to prevent C from approaching to 0. The corresponding weight $w_0$ is set to 0.1
\textbf{Bases Balance Loss}
\xwjrevise{
The final depth $D$ equals to $\sum_{i=1}^{N}{w_i B_i}$, where $B_i$ stands for $i$th basis. For given $K$ sparse points, we can estimate weight $W = [w_1 w_2 \dots w_N]^{\top}$ of bases by solving the following least square optimization problem:}

\begin{equation}
\label{equa:lsf}
    \arg \min_W  \sum_{j=1}^{K}( s_j - \sum_{i=1}^{N} w_i B_i(x_j)) ,
\end{equation}
\xwjrevise{
where $s_j$ is the depth of sparse point $j$, and $x_j$ is the corresponding 2d coordinate in image, $S$ is $[s_1 s_2 \dots s_K]^{\top}$. $B$ is defined as follows:}
\begin{equation}
B = \begin{bmatrix}
B_{1}(x_1)  & B_{2}(x_1) & \dots & B_{N}(x_1) \\
B_{1}(x_2)  & B_{2}(x_2) & \dots  & B_{N}(x_2) \\
\vdots   & \vdots  & \ddots   & \vdots \\
B_{1}(x_K) & B_{2}(x_K) & \dots  & B_{N}(x_K)
\end{bmatrix}.
\end{equation}
\color{black}

Then, we can rewrite the linear problem as follows:
 


\begin{equation}
\label{equa:origin_lsf}
(B^TB)W =B^T S .
\end{equation}

The condition number can be defined as follows:
 
\begin{equation}
\label{equa:cond}
cond = \frac{\lambda_{max}(B^T B)}{\lambda_{min}(B^T B)} ,
\end{equation}
 where $\lambda_{max}$, $\lambda_{min}$ is the max and min eigenvalue of the matrix.

When the bases exhibit significant imbalance, the condition number grows large and the problem becomes ill-posed. Consequently, even slight noise in the input sparse depth or the corresponding point of the predicted bases can result in substantial deviations in the weights, leading to erroneous outcomes. Hence, we incorporate the condition number into our loss function to mitigate such effects. This not only eliminates the interference of outlier values during training but also enables us to supervise the depth balance. According to~\cite{boeddeker2017computation}, the eigenvalue is differentiable; Therefore the condition number is also differentiable. We design the bases balance loss based on condition number, which is defined as follows:

 \begin{equation}
\label{equa:loss_cond}
    L_{b} = \log(\lambda_{max}(B^T B)) - \log(\lambda_{min}(B^T B)) .
\end{equation} 

 \textbf{Bases Balance Training} 
 During the depth prediction process, sparse depth is utilized in two steps: 1) the network combines the input image and sparse depth to predict the depth bases, and 2) the least squares module utilizes the predicted depth bases and sparse depth to solve for the weights of the depth bases. To prevent the network from overfitting to information near the sparse depth and causing an imbalance within the depth bases, we have chosen distinct sparse depth points from pre-extracted FAST corners during the training process for each of the two steps.


 



\section{Depth Weight Optimization}
\begin{figure}[!tb]
 \centering 
 \includegraphics[width=\columnwidth]{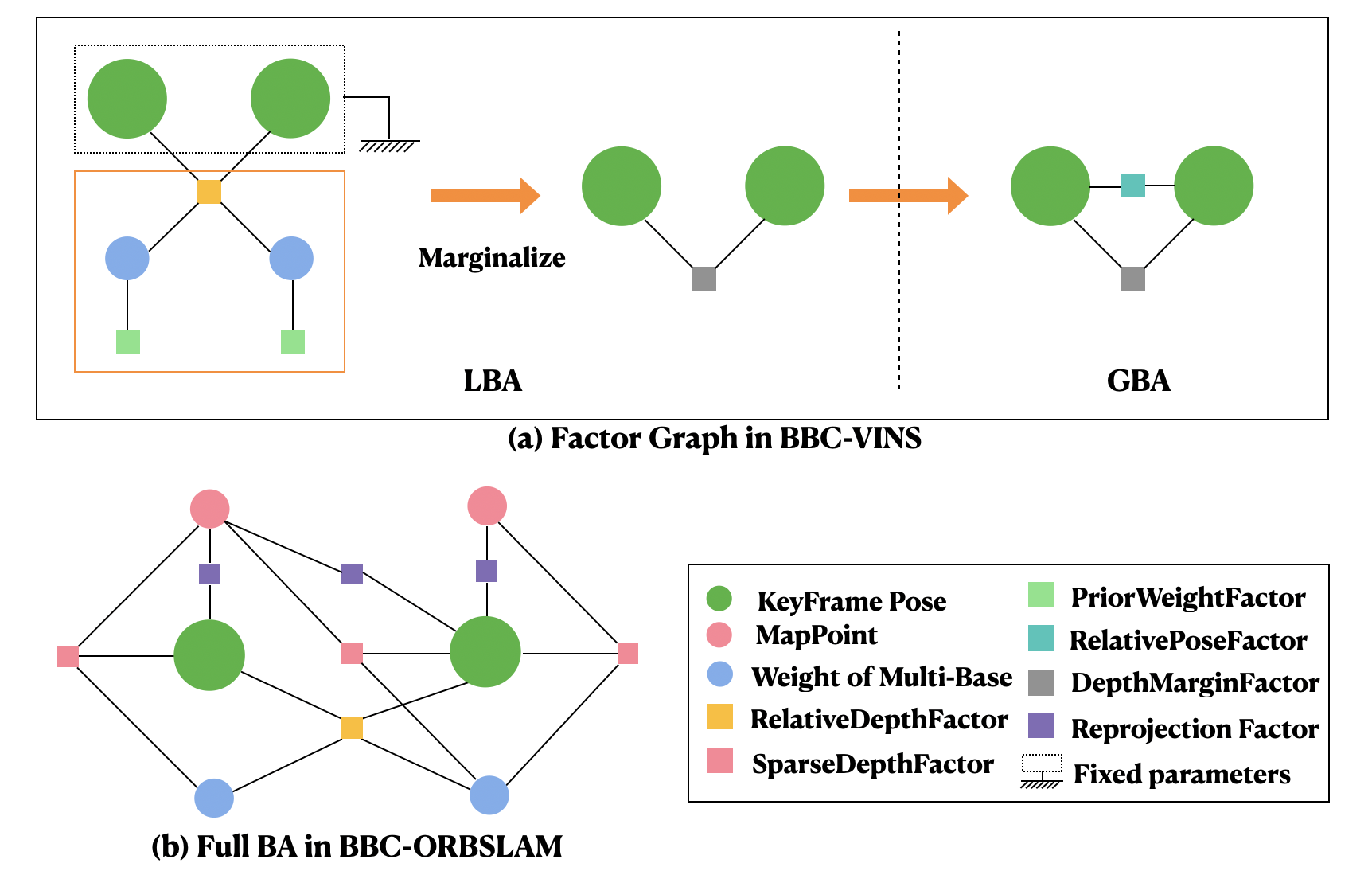}
 \caption{BA Factor Graph. (a) The factor graph of BBC-VINS. (b) The factor graph of BBC-ORB.}
 \label{fig:factor_graph}
\end{figure}

\xwjrevise{In this section, we introduce a set of factors based on the multi-basis depth representation. When integrated with a specific SLAM system, these factors can be flexibly combined to ensure precision and robustness in optimization. Additionally, the use of marginalization techniques facilitates an accelerated optimization process. By adding these factors into a traditional optimizer, we can establish constraints between the predicted depth and other variables for simultaneous optimization. Examples of the combination of various factors are illustrated in \autoref{fig:factor_graph}}. 
For convenience explanation, we pre-defined some notations and symbols here. $R_i$ and $t_i$ is rotation matrix and translation matrix of frame $i$ respectively. $w_i$ is the weight of bases of frame $i$, the dimention of $w_i$ is $n \times 1$, $n$ is the num of bases. $D_i$ are the depth bases of image $i$, $D_i(p)$ is bases values in point $p$ at image $i$, the dimension of $D_i(p)$ is equal to the dimension of weight $w_i$. $X_{w_k}$ is the position of map point $k$. 
$\pi$ is the projection function to project 3D point to image. Specifically, we define $R_z$ as $(0,0,1)^T$. 



\label{sec:depth_factor}
\subsection{Sparse Depth Weight Factor}
The sparse depth weight factor aligns the final depth with the sparse map points. As map points are observed by multiple keyframes, this factor can provide constraints between multiple keyframes.
Specifically, if we fix the pose of the keyframe and the position of map points, the function of this factor is equivalent to that of the least squares method described in~\cite{qu2020depth}. The \xwjrevise{residual term for sparse depth weight factor} can be defined as follows:
\begin{equation}
\label{equa:sparse_depth_weight_factor}
  {E}_{D_{ik}} =  R_z^{\top} (R_i X_{w_k}+t_i) - D_i\left ( \pi  \left ( R_i X_{w_k}+t_i \right )  \right ) ^ {\top} w_i .
\end{equation}

In light of the potential for SLAM to optimize the sparse points used as input for the BBC-Net, we did not employ the least squares solver described in~\cite{qu2020depth} during the online inference process. Instead, after the network predicts the depth bases, we obtain the latest map point coordinates and utilize this sparse depth weight factor to independently optimize the weight of the depth bases. 

\subsection{Relative Depth Weight Factor}
When enough overlap exists between frame $i$ and frame $j$, we can sample points in frame $i$ with a similar point sample method to~\cite{engel2017direct}. For sample point $p$ in frame $i$, we can get its corresponding depth $D_i^T(X_p)w_i$. Then, we can project this point to frame $j$ as
\begin{equation}
\label{equa:relative_depth_weight_factor_1}
  {X}_{c_j} = R_j(R_i^{\top} (\pi^{-1}(X_p) * D_i^{\top}\left(X_p\right)w_i-t_i))+t_j.
\end{equation}
So, the \xwjrevise{residual term for} relative depth \xwjrevise{weight factor} can be defined as follows:
\begin{equation}
\label{equa:relative_depth_weight_factor_2}
  {E}_{R_{ijp}} =  R_z^{\top} X_{c_j} - D_j^{\top}\left(\pi\left( X_{c_j} \right) \right) w_{j}.
\end{equation}
While the sparse depth weight factor \xwjrevise{imposes constraints across} multiple frames, \xwjrevise{it is based on the} map point \xwjrevise{ distributed as fast} corners \xwjrevise{limiting its influence on depth consistency. In contrast, the relative depth factor, which involves more evenly distributed sample points, plays a more pivotal role in achieving global depth consistency.}
\subsection{Prior Weight Factor}
The definition of the prior factor is relatively simple. The \xwjrevise{residual term} can be defined as follows:
\begin{equation}
\label{equa:prior_weight_factor}
  {E}_{P_{i}} =  \hat{w_i} - w_i,
\end{equation}
where $\hat{w_i}$ is the result of $w_i$ in last optimization iteration.

Under conditions of insufficient constraints, optimization divergence may occur. 
In fact, we consider it abnormal for the depth to undergo significant changes before and after optimization. Therefore, adding a prior factor is logical and can improve the robustness of the optimization.

\subsection{Weight Margininalization}
Generating a large number of sampling points that are added as observations to the optimizer can significantly reduce its efficiency. To improve the efficiency of the optimizer, we can use marginalization. When the oldest frame slides out of the LBA window, we marginalize the relative depth weight factors and prior weight factors related to the oldest frame and save these marginalized information separately. During GBA optimization, this marginalized information can be directly added to the optimizer. Similar to the method in~\cite{qin2018vins}, marginalization is achieved through Schur complement. 
Marginalizing the weights of the depth bases avoids redundant observations and significantly improves the optimizer's efficiency.




 
\section{SLAM Framework}
\label{sec:couple}

\xwjrevise{In this section, we provided concrete examples illustrating how to integrate the network and combine different factors into specific traditional SLAM frameworks.}

\subsection{BBC-VINS}
\label{sec:BBC-VINS}

\begin{figure}[!tb]
 \centering 
 \includegraphics[width=\columnwidth]{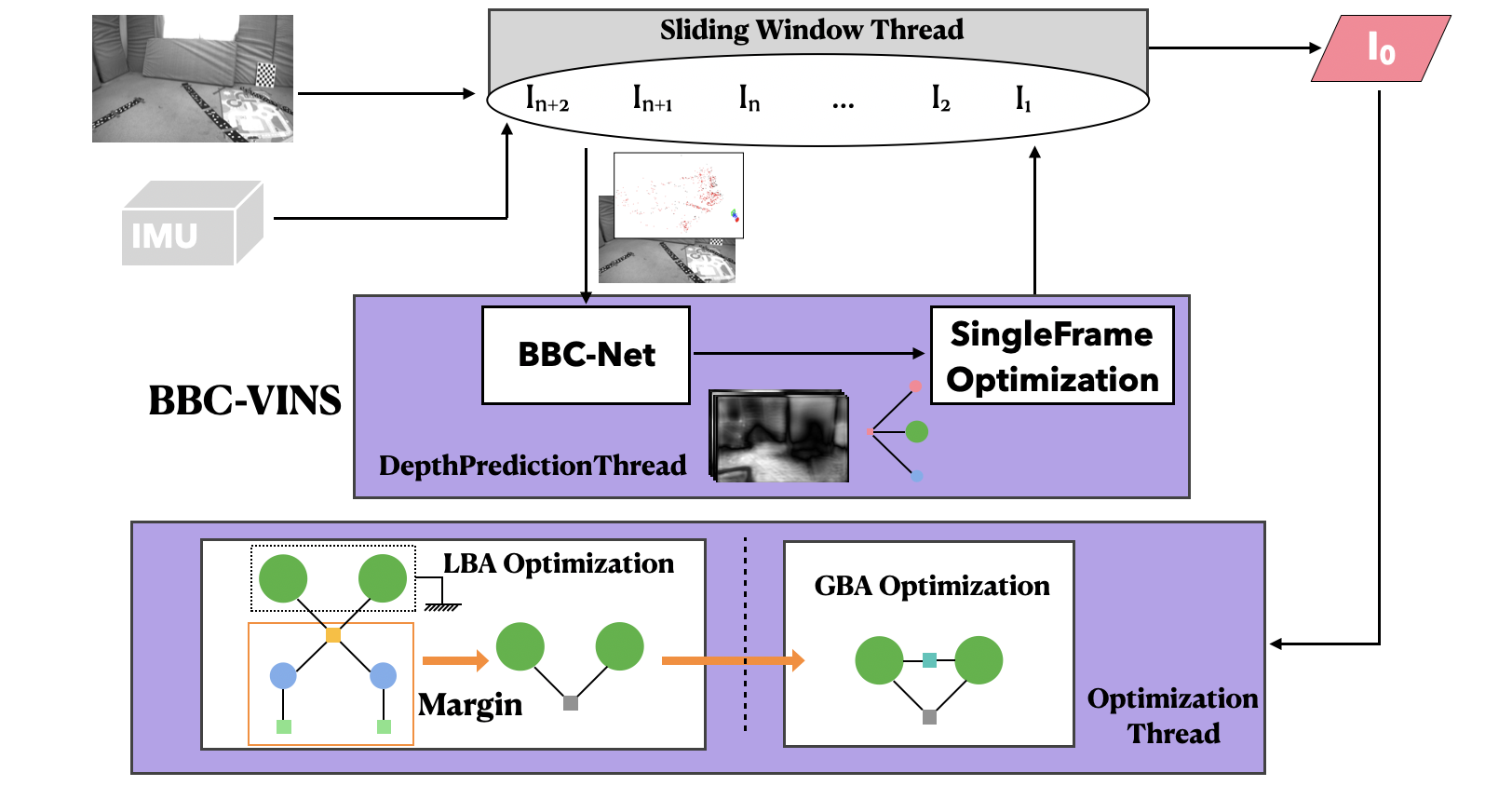}
 \caption{The architecture of BBC-VINS. In BBC-VINS, we employ a two-stage optimization strategy. In LBA optimization, the camera poses are fixed, and only the weights of the depth bases are optimized. On the other hand, camera poses are optimized in GBA optimization.}
 \label{fig:BBC-VINS}
\end{figure}
The architecture of BBC-VINS is shown in \autoref{fig:BBC-VINS}. 
In this paper, the BBC-VINS is developed based on our self-implemented VINS system called BVINS. BVINS follows the architecture of~\cite{qin2018vins} but replaces the sliding window optimization with \cite{wu2015square}. The length of the sliding window is set to 12. Moreover, BVINS utilizes the KLT optical flow tracker~\cite{shi1994good}, similar to~\cite{qin2018vins}, to track the features of FAST corners~\cite{rosten_2006_machine}. To ensure a minimum number of features (100-120) in each image, we supplement the FAST corners in the most recent frame for further visual tracking when the number of successfully tracked features is insufficient. When the frame is still in the sliding window, we package the image and sparse depth and send it to a separate thread for depth-based inference, which will not affect feature tracking and the efficiency of the sliding window. As the number of map points that can be tracked in a single frame is usually insufficient,  we project all points in the sliding window onto the same frame, retain the points still in the viewing range, and use them as sparse depth input to the BBC-Net.

 The main modification made in this work involves incorporating depth weight optimization into the backend optimization process. In a traditional pose graph backend, the optimization process only runs when a loop closure is detected. Fortunately, in BBC-VINS, the use of relative depth provides additional pose constraints, allowing for pose optimization even in the absence of loop closure. To ensure both efficiency and robustness, we have developed a two-stage optimization strategy involving LBA and GBA. During the LBA stage, only the weights of each depth base are optimized, while during the GBA stage, only the poses are optimized. 

\subsection{BBC-ORBSLAM}
\label{sec:system}

\begin{figure}[!tb]
 \centering 
 \includegraphics[width=\columnwidth]{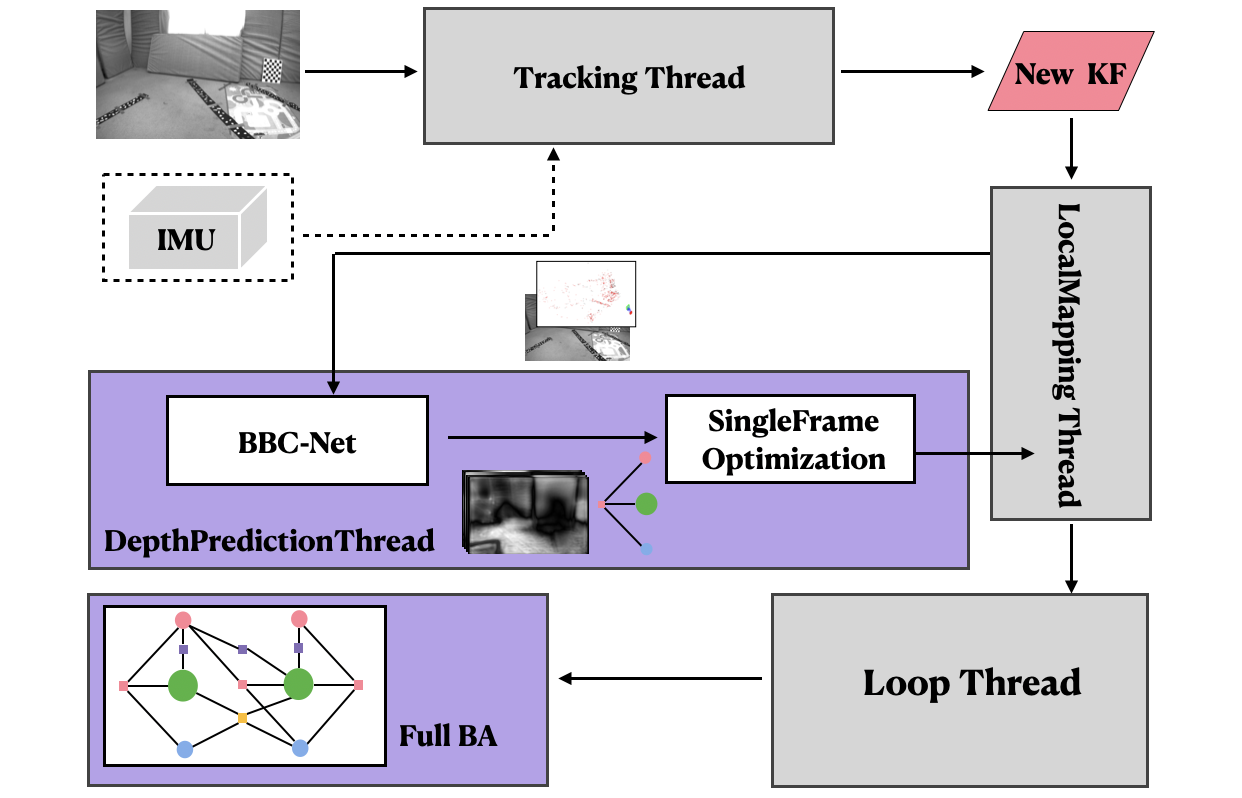}
 \caption{The architecture of BBC-ORBSLAM. We add the relative depth weight factor and the sparse depth weight factor to the optimizer, enabling joint optimization of camera poses, map points and the weights of depth bases.}
 \label{fig:BBC-ORB}
\end{figure}

The architecture of BBC-ORBSLAM is shown in ~\autoref{fig:BBC-ORB}.
Original ORB-SLAM3 consists of three threads: Tracking, LocalMapping, and Loop Closure. The keyframes are generated in the LocalMapping thread. Once a keyframe is generated, we will get all local map points from connected keyframes and then project local map points into the image of the current keyframe. After removing the point that cannot project into the image area or with invalid depth, the remaining sparse point combined with the gray image will be sent to BBC-Net, which runs in an independent thread. When a new candidate keyframe comes, the LocalMapping thread will try to get results from the BBC-Net thread. To solve the local inconsistency caused by multi-thread, we will optimize the weight of each base when the result comes back from BBC-Net. Due to the high global accuracy of ORB-SLAM3, we do not need to optimize all the weights at a high frequency. So we add the relative depth factor into Global Bundle Adjustment only.


\section{Experiments}
\label{sec:Experiments}
\xwjrevise{In this section, we conducted evaluations on the predicted depth, trajectory of dense SLAM, and reconstruction quality. For predicted depth, we compare BBC-Net with state-of-the-art depth completion method, including KBNet~\cite{wong2021unsupervised}, ACMNet~\cite{zhao2021adaptive}, NLSPN~\cite{park2020non}, CompletionFormer~\cite{zhang2023completionformer}. For the trajectory and reconstruction results, we compare our method with CNN-SLAM~\cite{tateno2017cnn}, CodeMapping~\cite{matsuki2021codemapping}, Tandem~\cite{koestler2022tandem}, DeepFactors~\cite{czarnowski2020deepfactors}, GeoRefine~\cite{ji2022georefine}, iMap~\cite{sucar2021imap} and NICE-SLAM~\cite{zhu2022nice}.}
As the relative depth factors will not be activated without global bundle adjustment in BBC-ORBSLAM, the trajectories and depths used in the experiment are the results after dense global bundle adjustment unless otherwise specified.
Unlike Tandem~\cite{koestler2022tandem}, our method does not depend on TSDF, so we save the depth and poses and utilize TSDF Fusion alone to generate the mesh.
\xwjrevise{
By default, we employ a model, trained for integration with the SLAM system, on a subset of ScanNet. This subset is generated using the same method as CodeMapping. 
}
In each frame, we sample 125 fast keypoints as sparse depth and enhance the sparse depth by perturbing 5\% of the depth values. Additionally, we introduce 2 pixels of Gaussian noise to the point coordinates.
 We conduct these comparative experiments on a desktop PC with an Intel i7-8700 CPU (3.2GHz *12) with 24 GB RAM, and use a Nvidia GTX 1060 6GB GPU for inference.
Due to space constraints, we provide only partial results in the paper. For additional implementation details\xwjrevise{, such as experiment setup} and more experiment results, please refer to the supplementary materials.


\subsection{\xwjrevise{Evaluation of Depth and Reconstruction}}
\subsubsection{\xwjrevise{Depth Completion Network}}
\setlength{\tabcolsep}{1.50mm}
\begin{table}
\centering
\caption{
\xwjrevise{
Evaluation of depth completion methods on NYUv2. The result labeled \textit{\textbf{\cite{qu2020depth}-O}} is extracted from the original paper, while \textit{\textbf{\cite{qu2020depth}-R}} represents our reimplementation of the baseline. Relative to \textit{\textbf{\cite{qu2020depth}-R}}, \textit{\textbf{Ours}} add the base balance loss.
}
}
\label{table:depth_completion_net}
\scalebox{0.92}[1.0]{
\begin{tabular}{c|c|c|c|c|c|c|c|c} 
\hline
    & \multicolumn{4}{c|}{random sparse} & \multicolumn{4}{c}{fast sparse} \\
\cline{2-9}
    & \multicolumn{2}{c|}{w/o noise} & \multicolumn{2}{c|}{w noise} & \multicolumn{2}{c|}{w/o noise} & \multicolumn{2}{c}{w noise} \\
\cline{2-9}
    & rmse $\downarrow$ & $\delta_{1}$ $\uparrow$ & rmse$\downarrow$ & $\delta_{1}$$\uparrow$ & rmse$\downarrow$ & $\delta_{1}$$\uparrow$ & rmse$\downarrow$ & $\delta_{1}$$\uparrow$ \\
\cline{1-9}
KBNet & 0.170 &0.986 & 1.752 &0.101 & 2.69 & 0.075& 2.691 & 0.072\\
\hline
ACMNet & 0.111 & 0.992 & 0.154 & 0.991 & 0.230 & 0.960 & 0.276 & 0.946 \\
\hline
NLSPN & 0.104 & 0.991 & 0.145 & 0.992 & 0.361 & 0.840 & 0.385 & 0.817 \\
\hline
CF\cite{zhang2023completionformer} & \textbf{0.097} & \textbf{0.995} & \textbf{0.139} & \textbf{0.993} & 0.223 & 0.966 & 0.270 & 0.956 \\
\hline
\textit{\textbf{Ours}} & 0.103 & 0.994 & 0.147 & 0.992 & \textbf{0.210} & \textbf{0.972} & \textbf{0.263} & \textbf{0.962} \\
\hline
\textit{\textbf{\cite{qu2020depth}-O}} & 0.134 & 0.993 & - & - & - & - & - & - \\
\hline
\textit{\textbf{\cite{qu2020depth}-R}} & 0.103 & 0.994 & 0.154 & 0.991 & 0.232 & 0.967 & 0.279 & 0.953 \\
\hline
\end{tabular}
}
\end{table}
\xwjrevise{
Considering variations in experimental setups among previous methods, we ensured a fair comparison by retraining all networks on the NYU dataset with identical dataset splits and training configurations.
From \autoref{table:depth_completion_net}, we can conclude the following conclusion: 1) \textit{\textbf{\cite{qu2020depth}-R}} significantly outperforms the origin \textit{\textbf{\cite{qu2020depth}-O}}. 2) Comparing \textit{\textbf{\cite{qu2020depth}-R}} and \textbf{Ours}, base balance loss can significantly enhance the network's robustness against fast corner distributions and sparse depth noise. 3) \textit{\textbf{Ours}} achieves slightly lower accuracy than NLSPN and CompletionFormer in the case of random sparse points, but outperforms them with fast corner points distribution. The observed discrepancy stems from the network's ability to improve accuracy through overfitting when dealing with randomly distributed sparse depth. In contrast, achieving accuracy through overfitting becomes challenging, and may even lead to adverse effects(see NLSPN), when dealing with sparse depths distributed as fast corners. The base balance loss helps mitigate this overfitting, which is why our method performs better when the sparse depth follows the distribution of fast corner points.}

\subsubsection{\xwjrevise{SLAM System with Predict Depth}}
\label{sec:ex_dense_mapping}
\xwjrevise {
 The versatility of ORB-SLAM, which supports multiple sensor inputs, enables our BBC-ORBSLAM to seamlessly handle various sensor inputs. For simplicity, we adopt the notations -V, -VI, and -VD to denote visual-only, visual-inertial, and RGBD modes, respectively.}
We first evaluate our method on the widely used EuRoC dataset~\cite{burri2016euroc}.  
Trajectory and depth evaluations for each frame are presented in \autoref{table:euroc_depth} and \autoref{table:euroc_traj}. 
\xwjrevise{As ~\cite{matsuki2021codemapping},} we generate ground truth depth for each frame by rendering LiDAR data with the ground truth pose.
\xwjrevise{The ground truth used to evaluate Tandem's depth is obtained from their preprocessed Euroc dataset.}
Specifically, we use Sim(3) alignment to refine the trajectory and the scale of corresponding depths for monocular methods. 

\xwjrevise{As depicted in \autoref{table:euroc_depth} and \autoref{table:euroc_traj}, BBC-VINS and BBC-ORB-VI outperform other methods in visual-inertial mode, while BBC-ORB-V achieves results comparable to GeoRefine in visual-only mode. It's important to note that although both BBC-ORBSLAM and GeoRefine are built upon ORB-SLAM3, BBC-ORBSLAM maintains the standard tracking module without customized modifications. In contrast, GeoRefine incorporates Raft-Flow in the front-end tracking module (a modification point independent of the depth completion network), resulting in enhanced tracking accuracy and point cloud quality.
Given the flexibility of our method as a plugin, we anticipate that its adaptation to a higher-precision SLAM system, such as GeoRefine, could yield even superior results. Comparing CodeVIO and CodeMapping with BBC-VINS and BBC-ORBSLAM separately, it is evident that our method surpasses previous approaches in optimizing both depth and pose under comparable baseline SLAM conditions. Especially, the accuracy of dense depth generated by our method is significantly higher than the sparse depths produced by baseline SLAM. In contrast, CodeVIO and CodeMapping face challenges in achieving higher accuracy of final depth than the sparse depth generated by baseline SLAM. 
}

\setlength{\tabcolsep}{2.00mm}
\begin{table}
\centering
\caption{Evaluation on depth RMSE(m) on EuRoC. The best result is shown in bold, the second result is shown with underline. \xwjrevise{The sparse depth error of baseline SLAM is shown in gray}}
\label{table:euroc_depth}
\scalebox{1.0}[1.0]{
\begin{tabular}{ccccccc} 
\hline
         & V101                      & V102                               & V103                               & V201                               & V202                               & V203                                \\ 
\hline
DeepFactor~\cite{czarnowski2020deepfactors}\       & 1.05                      & 1.03                               & 0.94                               & 1.02                               & 1.53                               & \underline{1.25}                       \\
Tandem~\cite{koestler2022tandem}        & 0.439                     & 0.375                              & 0.858                              & 0.393                              & 0.559                              & 2.82                                \\
GeoRef-V~\cite{ji2022georefine}       & \textbf{0.241}            & \multicolumn{1}{l}{\underline{0.255}} & \multicolumn{1}{l}{\textbf{0.297}} & \multicolumn{1}{l}{\underline{0.258}} & \multicolumn{1}{l}{\textbf{0.208}} & \multicolumn{1}{l}{\textbf{0.231}}  \\
BBC-ORB-V  & \underline{0.255}                     & \textbf{0.163}                     & \underline{0.325}                     & \textbf{0.255}                     & \underline{0.355}                     & -                                   \\ 
\hline

GeoRef-VI~\cite{ji2022georefine}        & \multicolumn{1}{l}{0.241} & \multicolumn{1}{l}{0.251}          & \multicolumn{1}{l}{0.278}          & \multicolumn{1}{l}{0.258}          & \multicolumn{1}{l}{\underline{0.312}} & \multicolumn{1}{l}{0.220}           \\
CodeMap~\cite{matsuki2021codemapping}        & 0.381                     & 0.369                              & 0.407                              & 0.428                              & 0.655                              & 0.952                               \\
\textcolor{gray}{CodeMap~\cite{matsuki2021codemapping}-sp}        & \textcolor{gray}{0.283}                     & \textcolor{gray}{0.381}                              & \textcolor{gray}{0.405}                              & \textcolor{gray}{0.374}                              & \textcolor{gray}{0.617}                              & \textcolor{gray}{0.801}                               \\
BBC-ORB-VI & \underline{0.211}                     & \textbf{0.155}                     & \textbf{0.169}                     & \underline{0.243}                     & 0.326                              & \textbf{0.198}                      \\
\textcolor{gray}{BBC-ORB-VI-sp}        & \textcolor{gray}{0.494}                     & \textcolor{gray}{0.235}                              & \textcolor{gray}{0.443}                              & \textcolor{gray}{0.443}                              & \textcolor{gray}{0.760}                              & \textcolor{gray}{0.651}                               \\
CodeVIO   & 0.468            & 0.602                     & 0.687                     & 0.656                     & 0.777                     & 0.652   
\\
\textcolor{gray}{CodeVIO-sp}   & \textcolor{gray}{0.240}            & \textcolor{gray}{0.399}                     & \textcolor{gray}{0.463}                     & \textcolor{gray}{0.506}                     & \textcolor{gray}{0.572}                     & \textcolor{gray}{0.488}                      \\
BBC-VINS   & \textbf{0.186}            & \underline{0.164}                     & \underline{0.176}                     & \textbf{0.241}                     & \textbf{0.244}                     & \underline{0.209}                      \\
\textcolor{gray}{BBC-VINS-sp}   & \textcolor{gray}{0.511}            & \textcolor{gray}{0.489}                     & \textcolor{gray}{0.579}                     & \textcolor{gray}{0.486}                     & \textcolor{gray}{0.756}                     & \textcolor{gray}{0.681}                      \\
\hline
\end{tabular}
}
\end{table}

\begin{table}
\centering
\caption{Evaluation on trajectory on EuRoC with APE(m). The best result is shown in bold, the second result is shown with underlined. \xwjrevise{The results of baseline SLAM are shown in gray}}
\label{table:euroc_traj}
\begin{tabular}{ccccccc} 
\hline
                   & V101           & V102           & V103           & V201           & V202           & V203            \\ 
\hline
DeepFactor\cite{czarnowski2020deepfactors}                 & 1.520          & 0.680          & 0.900          & 0.880          & 1.910          & 1.020           \\
Tandem\cite{koestler2022tandem}             & 0.100          & 0.185          & 0.187          & 0.106          & 0.101          & 0.602           \\
GeoRef-V\cite{ji2022georefine}          & \textbf{0.032} & \textbf{0.010} & \textbf{0.022} & \underline{0.019} & \textbf{0.011} & \textbf{0.025}  \\
BBC-ORB-V            & \underline{0.035}          & \underline{0.012} & \underline{0.076} & \textbf{0.016} & \underline{0.036} & -               \\ 
\hline
CodeVIO\cite{zuo2021codevio}            & 0.054          & 0.071          & \underline{0.068} & 0.097          & \underline{0.061} & 0.275           \\
\textcolor{gray}{OpenVINS\cite{Geneva2020ICRA}}           & \textcolor{gray}{0.056}          & \textcolor{gray}{0.072}          & \textcolor{gray}{0.069}          & \textcolor{gray}{0.098}          & \textcolor{gray}{0.061}          & \textcolor{gray}{0.286}           \\
BBC-VINS             & \underline{0.039} & \underline{0.069} & 0.096          & \underline{0.043} & 0.108          & \underline{0.147}  \\
\textcolor{gray}{BVINS}              & \textcolor{gray}{0.042}          & \textcolor{gray}{0.072}          & \textcolor{gray}{0.107}          & \textcolor{gray}{0.045}          & \textcolor{gray}{0.113}          & \textcolor{gray}{0.154}           \\
BBC-ORB-VI           & \textbf{0.033} & \textbf{0.011} & \textbf{0.013} & \textbf{0.018} & \textbf{0.013} & \textbf{0.018}  \\ 
\hline
\end{tabular}
\end{table}

\begin{figure}[!htb]
\centering

\subfloat[Tandem-V201] {
\includegraphics[width=0.45\columnwidth]{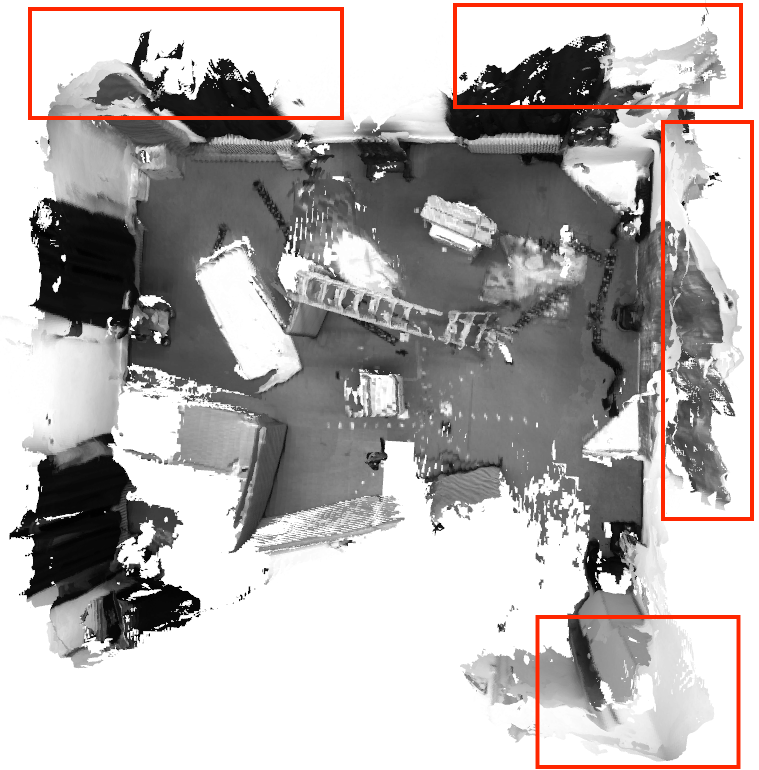}
} 
\subfloat[CompletionFormer-V201] {
\includegraphics[width=0.45\columnwidth]{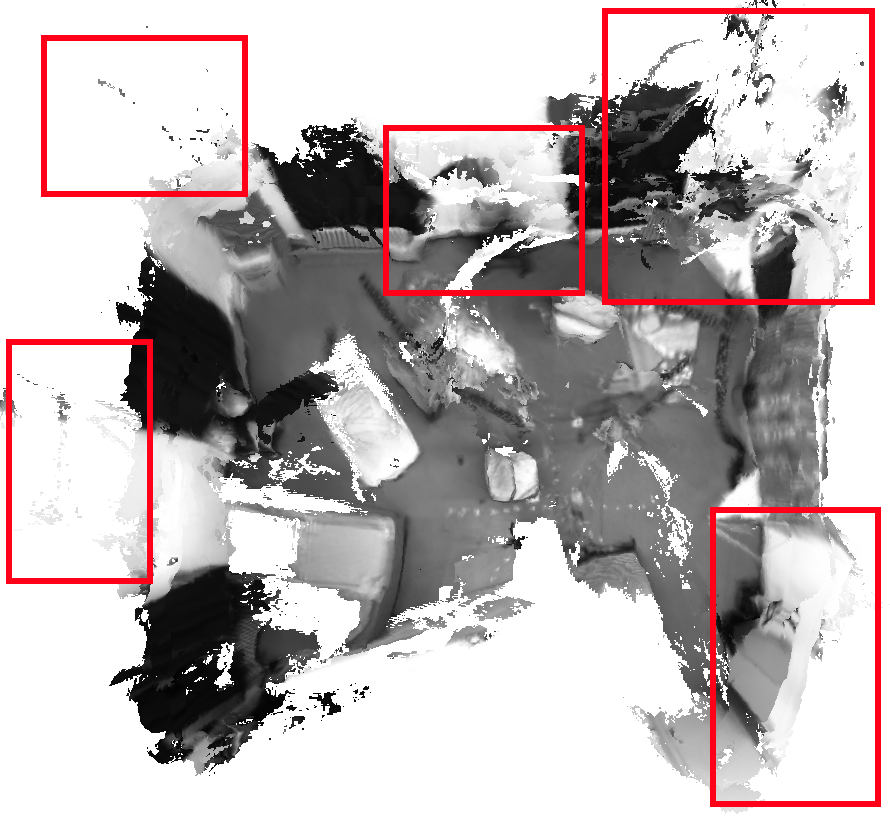}
} \\
\subfloat[BBC-VINS-V201] {
\includegraphics[width=0.45\columnwidth]{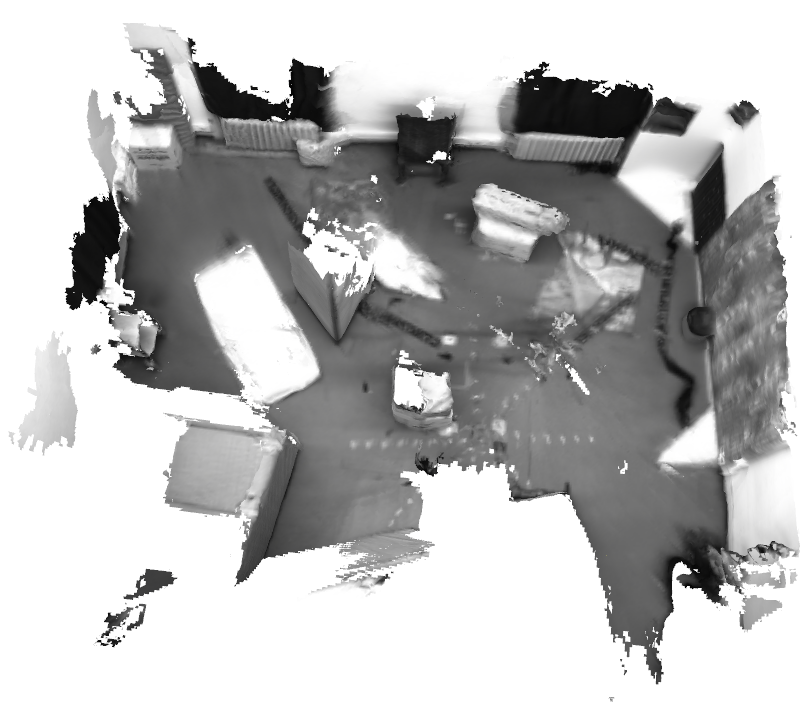}
} 
\subfloat[BBC-ORB(VI)-V201] {
\includegraphics[width=0.45\columnwidth]{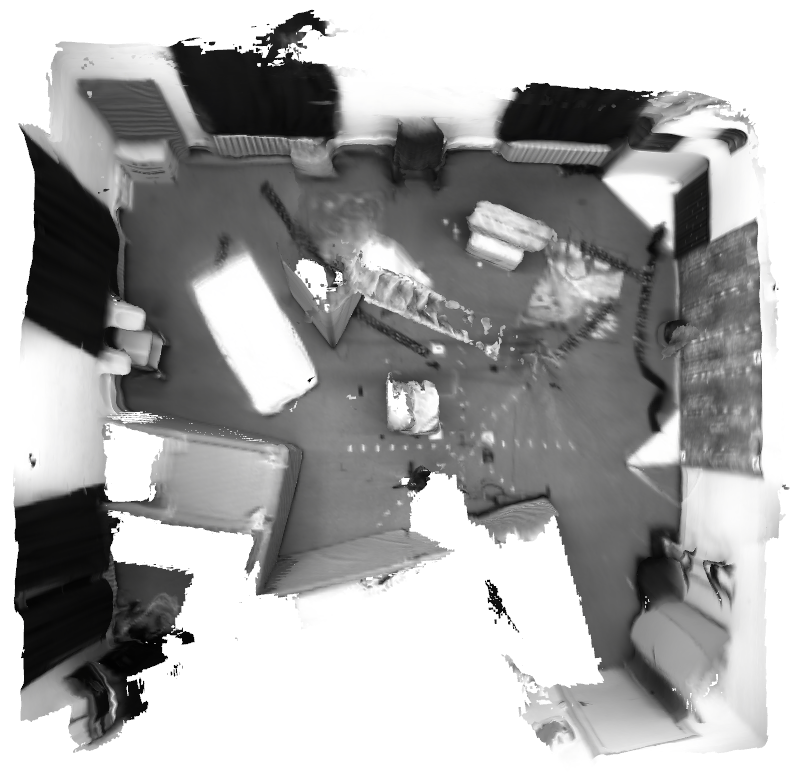}
} \\

\caption{ 3D Models generated by different methods in EuRoC. The red box highlights the portion of the reconstructed mesh with significant errors.}  
\label{fig:euroc_dense_mesh}
\end{figure}

\xwjrevise{We performed a qualitative assessment of mesh quality, as depicted in \autoref{fig:euroc_dense_mesh}. 
In addition to comparing with Tandem, we retrained CompletionFormer on ScanNet using identical settings to our network and replaced BBC-Net with it in BBC-VINS. Figure \ref{fig:euroc_dense_mesh}(b) and Figure \ref{fig:euroc_dense_mesh}(c) use the same images, sparse depths and poses. Visual analysis of the results unequivocally demonstrates that meshes generated by Tandem and CompletionFormer exhibit more noise compared to our method. The results from CompletionFormer also indicate that even state-of-the-art depth completion networks struggle to achieve good reconstruction results when integrated with SLAM systems, lacking necessary customizations and multi-frame depth optimization.}

\setlength{\tabcolsep}{0.2mm}
\begin{table}
\centering
\caption{Evaluation of the percentage of correct pixels \xwjrevise{for which pixels fall within 10\% of the groundtruth depth} on ICL-NUIM datasets. \cite{koestler2022tandem} was trained on Replica datasets. The (*) represents the model trained on ScanNet.
}
\label{table:icl_depth}
\scalebox{0.92}[0.9]{
\begin{tabular}{cccccccc} 
\hline
  & \begin{tabular}{c}\xwjminor{CNN \cite{tateno2017cnn}}\end{tabular}
  & \begin{tabular}{c}\xwjminor{DF \cite{czarnowski2020deepfactors}}\end{tabular}
  & \begin{tabular}{c}\xwjminor{Tdem \cite{koestler2022tandem}}*\end{tabular}
  & \begin{tabular}{c}\xwjminor{Tdem \cite{koestler2022tandem}}\end{tabular}
  & \begin{tabular}{c}\xwjminor{\textit{\cite{qu2020depth}-R}\textit{-VD}}\end{tabular}
  & \begin{tabular}{c} \xwjminor{\textit{{Ours}} \textit{{-V}}}\end{tabular} 
  & \begin{tabular}{c} \xwjminor{\textit{{Ours}} \textit{{-VD}}}\end{tabular}  \\ 
\hline
\multicolumn{1}{c}{office0} & 19.41 & 30.17 & 52.34 & 84.04 & \xwjrevise{53.36} & 79.52 & \textbf{84.39} \\
\multicolumn{1}{c}{office1} & 29.15 & 20.16 & 61.83 & 91.18 & \xwjrevise{52.35} & 64.70 & \textbf{96.54} \\
\multicolumn{1}{c}{living0} & 12.84 & 20.44 & 86.42 & 97.00 & \xwjrevise{72.83} & 95.73 & \textbf{97.62} \\
\multicolumn{1}{c}{living1} & 13.03 & 20.86 & 71.35 & 90.62 & \xwjrevise{73.08} & 73.27 & \textbf{95.13} \\
\hline
\end{tabular}
}
\end{table}

\minorrevise
\setlength{\tabcolsep}{0.7mm}
\begin{table}[]
\centering
\caption{Evaluation of depth accuracy on TUM datasets.(seq1: fr3 long office household; seq2: fr3 nostructure texture near withloop; seq3:fr3 structure texture far)}
\label{table:tum}
\scalebox{0.9}[0.9]{
\begin{tabular}{cccccccccc}
\hline
 &
 & \begin{tabular}[c]{@{}c@{}}\xwjminor{CNN\cite{tateno2017cnn}}\end{tabular} 
 & \begin{tabular}[c]{@{}c@{}}\xwjminor{DF\cite{czarnowski2020deepfactors}}\end{tabular} 
 & \begin{tabular}[c]{@{}c@{}} \xwjminor{\cite{ji2022georefine}-V}\end{tabular} 
 & \begin{tabular}[c]{@{}c@{}} \xwjminor{\cite{ji2022georefine}-VD}\end{tabular} 
 & \begin{tabular}[c]{@{}c@{}} \xwjminor{\textit{\cite{qu2020depth}-R-VD}}\end{tabular} 
 & \begin{tabular}[c]{@{}c@{}}\xwjminor{\textit{Ours} \textit{-V}}\end{tabular} 
 & \begin{tabular}[c]{@{}c@{}}\xwjminor{\textit{Ours} \textit{-VD}}\end{tabular} \\ \hline
\multirow{2}{*}{seq1}   & $\delta_1$ & 12.48 & 29.33 & -     & -  & \xwjrevise{94.14} & 81.23 & \textbf{95.40} \\
                       & RMSE     & -     & -     & - & - & \xwjrevise{0.437} & 0.253 & \textbf{0.132}  \\ \hline
\multirow{2}{*}{seq2}  & $\delta_1$ & 24.08 & 16.92 & -     & - & \xwjrevise{95.50} & 92.90 & \textbf{99.55} \\
                       & RMSE     & -     & -     & - & - & \xwjrevise{0.045} & 0.057 & \textbf{0.027}  \\ \hline
\multirow{2}{*}{seq3} & $\delta_1$ & 27.40 & 51.85 & -     & - & \xwjrevise{97.62} & 88.25 & \textbf{99.05} \\
                      & RMSE     & -     & -  & 0.317 & 0.290  & \xwjrevise{0.275} & 0.303 & \textbf{0.064}  \\ \hline
\end{tabular}
}
\end{table}
\color{black}

\xwjrevise{We extended our evaluation to include the ICL-NUIM~\cite{handa2014benchmark} and TUM~\cite{sturm2012benchmark} datasets, both widely utilized RGBD datasets.}
\begin{figure}[!htb]
\centering
\subfloat[RGB Mode] {
\includegraphics[width=0.46\columnwidth]{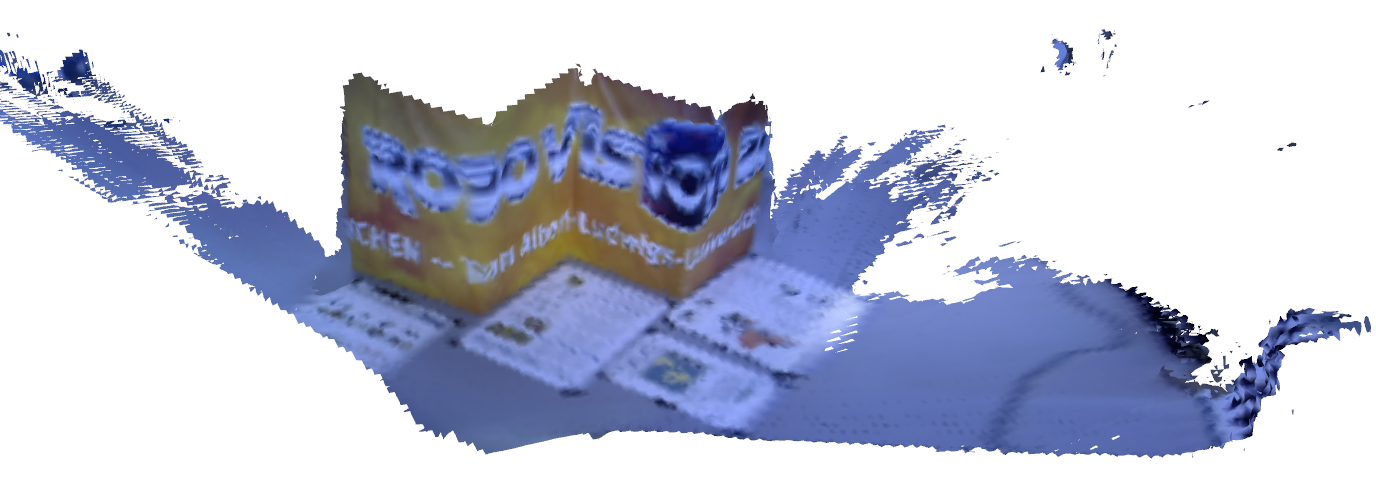}
}
\subfloat[RGBD Mode] {
\includegraphics[width=0.46\columnwidth]{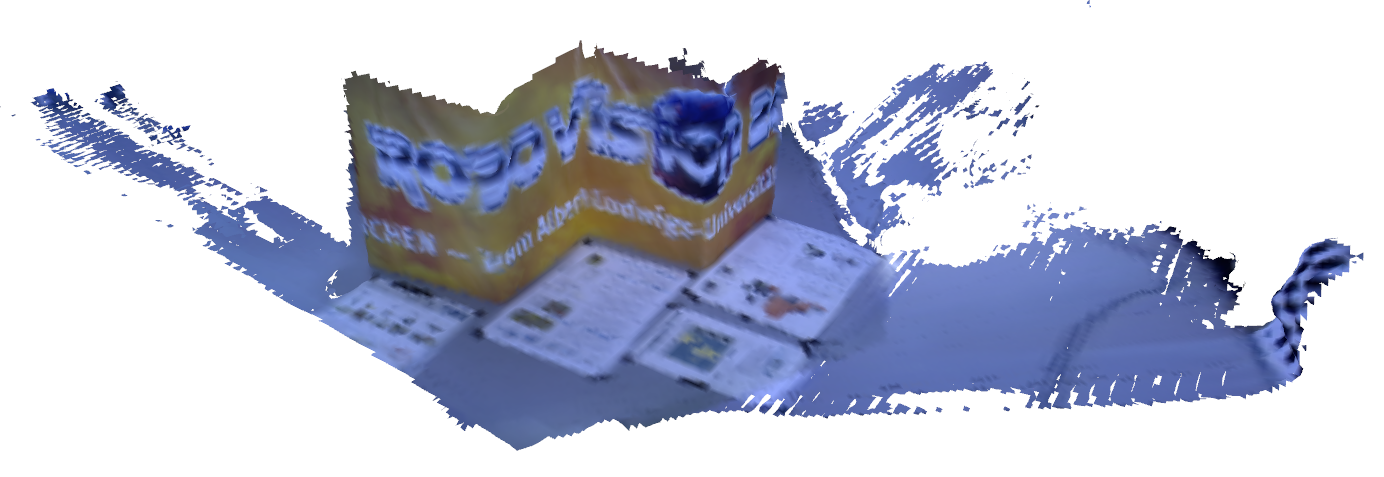}
}
\caption{ (a) The mesh generated by BBC-ORB-V. (b) The mesh generated by BBC-ORB-VD. }  
\label{fig:tum_dense_mesh}
\end{figure}
When running BBC-ORB-VD, similar to~\cite{matsuki2021codemapping}, we only use depth for tracking, while relying on the 3D points generated by SLAM itself for depth prediction and optimization.
\autoref{table:icl_depth} and \autoref{table:tum} present our results on the ICL and TUM datasets, respectively. 
\xwjrevise{\textit{\cite{qu2020depth}-R-VD} directly integrates \textit{\cite{qu2020depth}-R} network with the SLAM system. Comparing \textit{\cite{qu2020depth}-R-VD} and \textit{Ours-VD} highlights the necessity of our improvements over \cite{qu2020depth}, as they significantly enhance the performance when integrating with SLAM systems.}
The performance of our BBC-ORB-V was hindered on several sequences of the ICL dataset due to the presence of textureless regions, which pose a significant challenge for monocular SLAM systems.
 \autoref{fig:tum_dense_mesh} indicates that for texture-rich sequence, BBC-ORB-V can achieve comparable results to BBC-ORB-VD.

\begin{figure*}[!htb]
\centering
\subfloat[] {
\includegraphics[width=0.36\columnwidth]{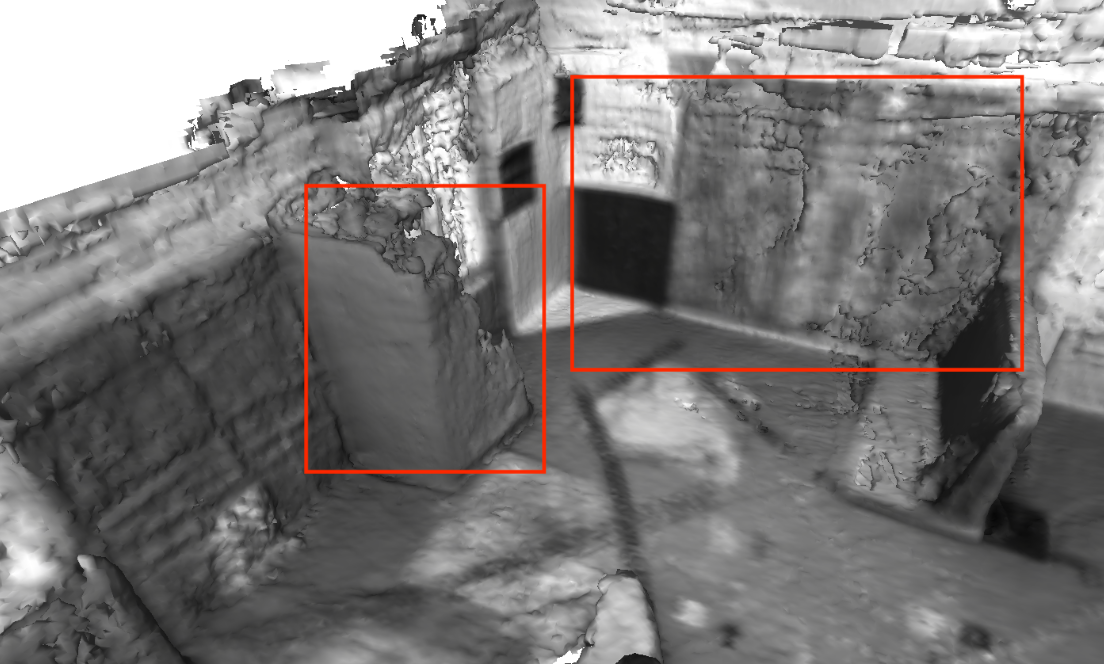}
}
\subfloat[] {
\includegraphics[width=0.36\columnwidth]{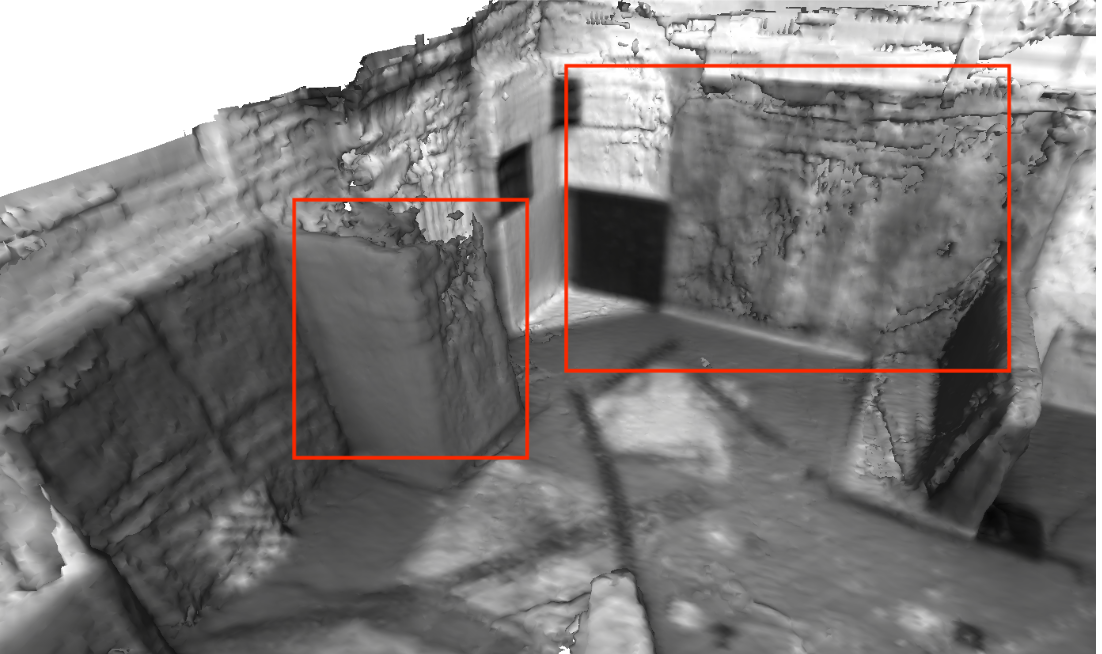}
}
\subfloat [] {
\includegraphics[width=0.36\columnwidth]{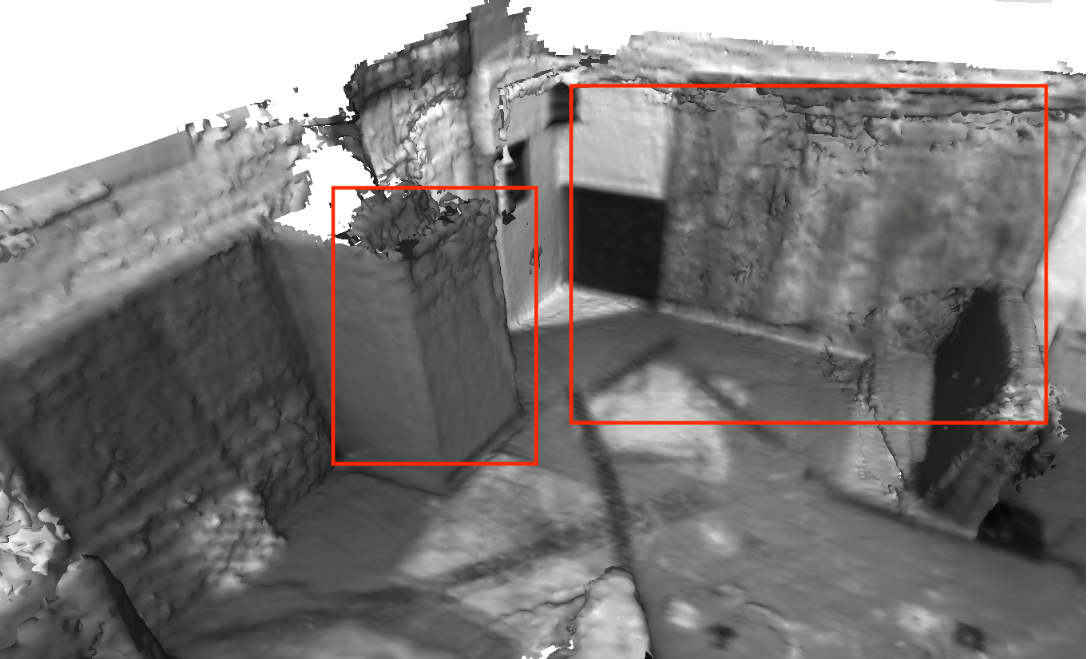}
}
\subfloat [] {
\includegraphics[width=0.36\columnwidth]{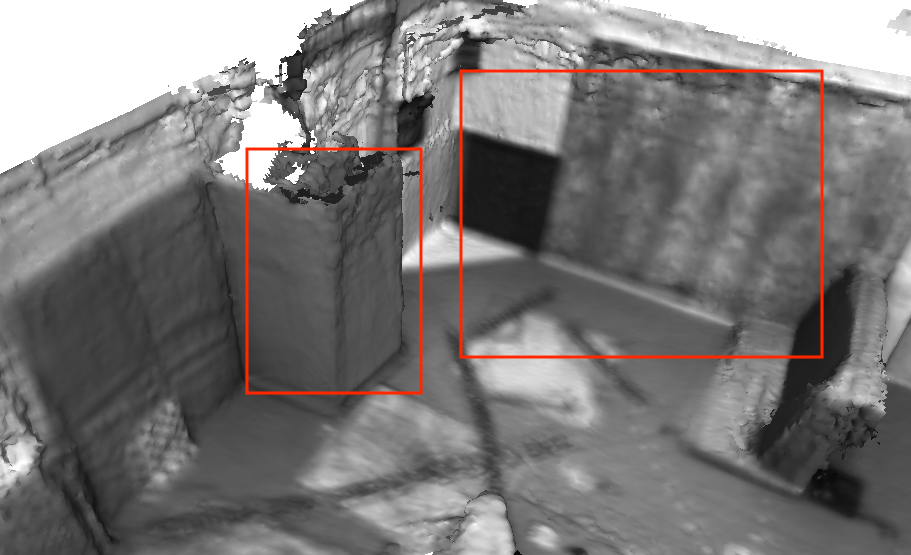}
}
\subfloat [] {
\includegraphics[width=0.36\columnwidth]{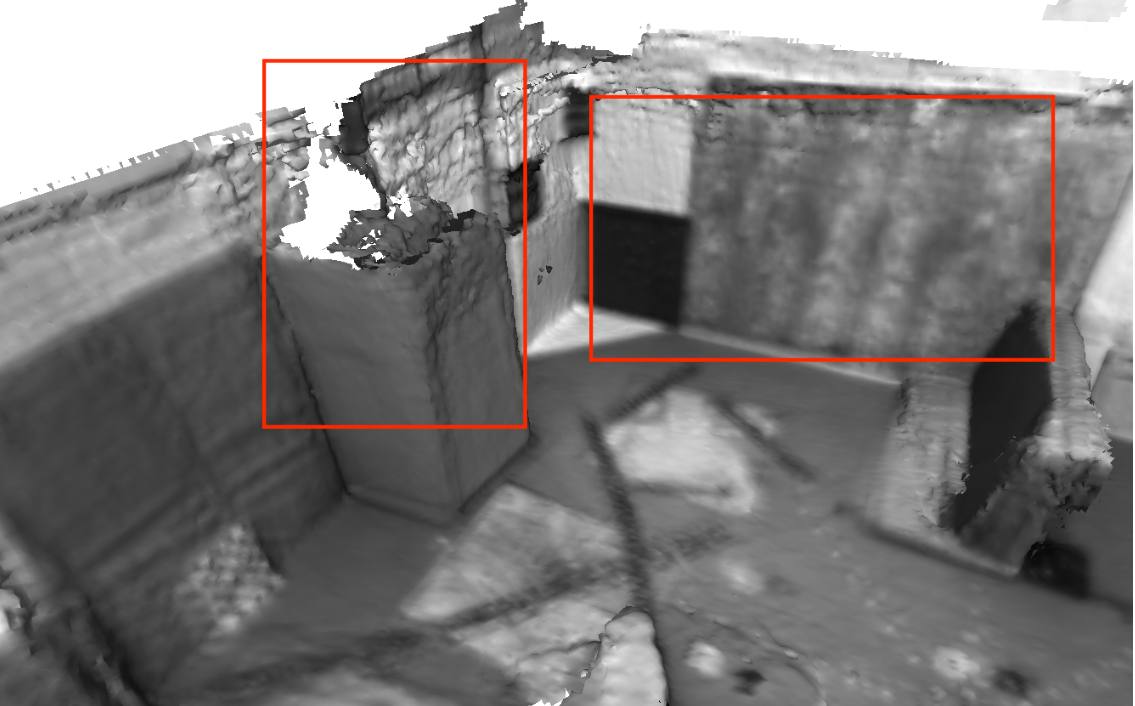}
}

\caption{
Meshes generated with BBC-VINS in V102. (a) The results of~\cite{qu2020depth}.  (b) The results of +\textbf{\textit{B}}. (c) The results of +(\textbf{\textit{B}}, \textbf{\textit{C}}).  (d) The results of +(\textbf{\textit{full}}, \textbf{\textit{M}}). (e) The results of +(\textbf{\textit{full}}).
}  
\label{fig:ablation_study_v12}
\end{figure*}

\begin{figure*}[!htb]
\centering

\subfloat[] {

\includegraphics[width=0.45\columnwidth]{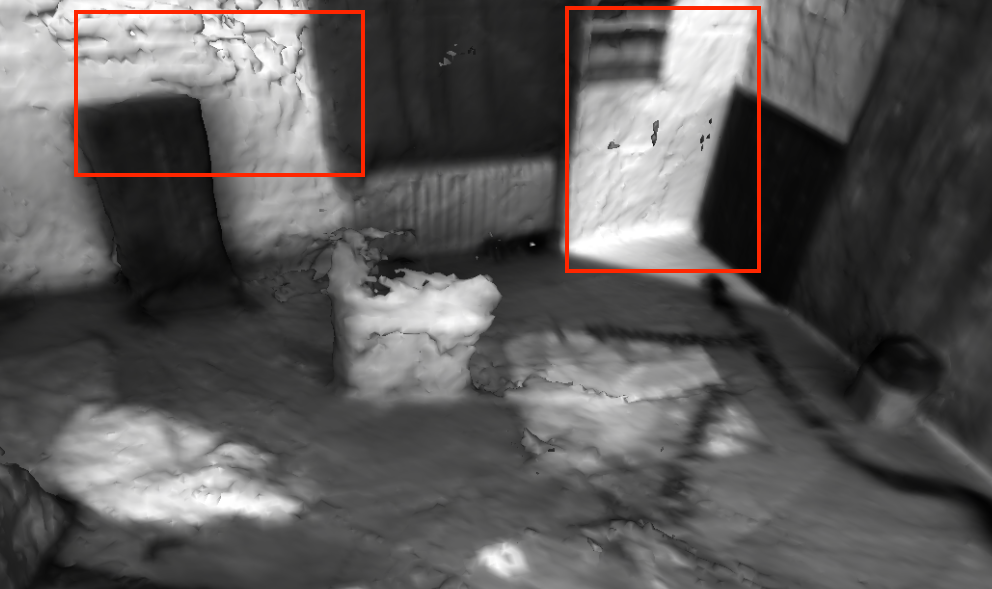}
}
\subfloat[] {
\includegraphics[width=0.45\columnwidth]{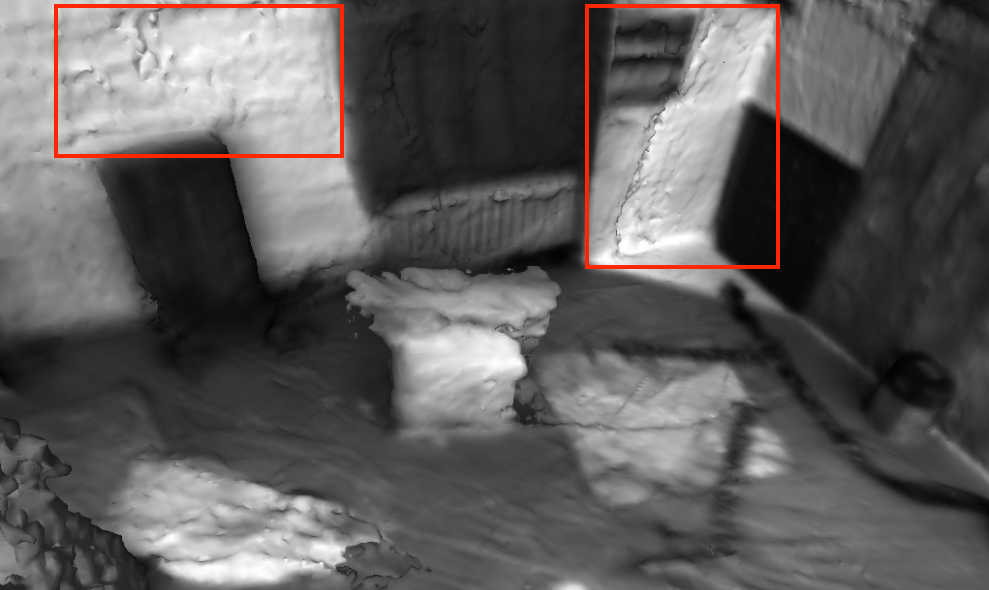}
}
\subfloat [] {
\includegraphics[width=0.45\columnwidth]{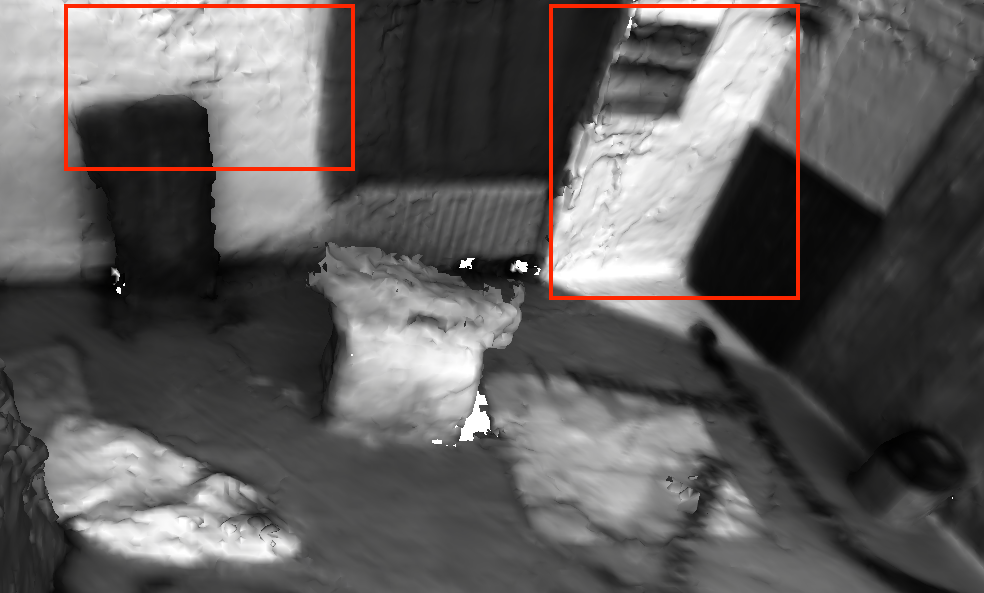}
}
\subfloat [] {
\includegraphics[width=0.45\columnwidth]{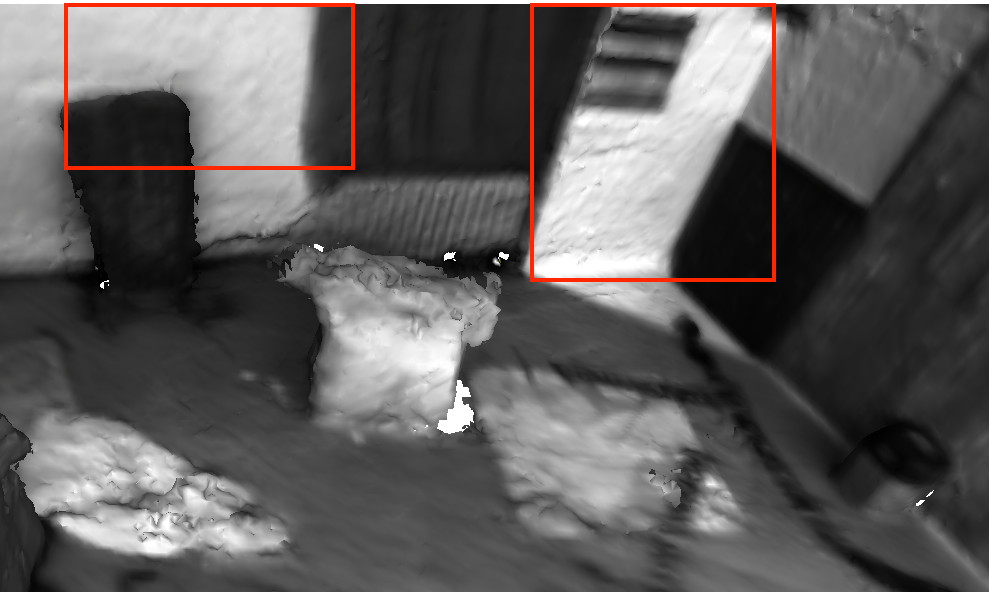}
}

\caption{Meshes generated with BBC-ORB-VI in V203. 
(a) The result of~\cite{qu2020depth}.  (b) The result of +(\textbf{\textit{B}}). (c) The result of +(\textbf{\textit{B}}, \textbf{\textit{C}}).  (d) The result of +(\textbf{\textit{full}}).
}  
\label{fig:ablation_study_v23_orbvi}
\end{figure*}

\subsubsection{End-to-End Implicit Dense SLAM}

\setlength{\tabcolsep}{1.30mm}
\begin{table}
\centering
\caption{\xwjrevise{Reconstruction result for the Replica Dataset(average over 7 scenes except office1). 
Ours-VD exclusively employs fast corners, while Ours-VD* utilizes a combination of fast corners and random sparse points, leveraging a retrained model.
}
}
\label{table:replica}
\scalebox{0.96}[1.0]{
\begin{tabular}{c|c|c|c|c|c} 
\hline
    & TSDF-Fusion & iMap & NICE-SLAM & Ours-VD & Ours-VD* \\
\hline
Acc.$\downarrow$ & 3.559 & 4.533 & 3.707 & 6.844 & 2.263  \\
\hline
Comp.$\downarrow$ & 6.104 & 5.606 & 3.851 & 7.686 & 4.210  \\
\hline
Comp. Ratio$\uparrow$ & 75.54 & 78.97 & 84.19 & 56.65 & 82.34  \\
\hline
\end{tabular}
}
\end{table}

\xwjrevise{
The Replica dataset is a simulated dataset, exhibiting significant differences from the real-world dataset, Scannet, we used for training. These differences are particularly evident in the distribution of fast corners and patterns of depth variation.
To mitigate this distinction, we augmented the fast corners with approximately 350 additional random sparse points, bringing the sparse depth number to 500.
Subsequently, we retrain our network on Scannet to adapt to the distribution characteristics of a mixture of fast points and random sparse points. It is important to note that in \autoref{table:replica}, all methods, excluding ours, utilize the entire ground truth depth as input. In contrast, we only employed approximately 1\% of the depth values. 
Surprisingly, with the model retrained on the Scannet only and utilizing much fewer ground truth depth points compared to other methods, our method achieved results comparable to NICE-SLAM and significantly outperformed iMap.
Furthermore, our method exhibits greater efficiency compared to iMap and NICE-SLAM, as discussed in \autoref{sec:runtime_analyze}. The omission of the "office1" sequence from the statistics is attributed to ORB-SLAM3's failure to track on this specific sequence.}

\subsection{Ablation Study}
\label{sec:ablation_study}
\xwjrevise{To assess each module's individual contribution in our system, we conducted ablation experiments. Simplified symbols represent optimization configurations for clarity: \textbf{\textit{B}} for balanced bases, \textbf{\textit{C}} for confidence, and \textbf{\textit{full}} for balanced base, confidence, and BA optimization. Additionally, \textbf{\textit{M}} denotes the use of marginalization in optimization.}

\setlength{\tabcolsep}{2.mm}
\begin{table}
\centering
\caption{Evaluation of BBC-VINS on EuRoC. \textit{D} stand for depth error.
}
\label{table:ablation_BBC-VINS}
\scalebox{0.96}[1.0]{
\begin{tabular}{c|c|cccccc} 
\hline
    \multicolumn{2}{c}{} & V101 & V102 & V103 & V201 & V202 & V203 \\
\hline
\multirow{6}{*}{Depth} & \begin{tabular}{c} \textit{Sp} \end{tabular}  & 0.511 & 0.489 & 0.579 & 0.486 & 0.756 & 0.681     \\
\cline{2-8}
                      & \begin{tabular}{c} \textit{\textbf{\cite{qu2020depth}-R}}
                        \end{tabular}  & 0.367 & 0.359 & 0.383 & 0.451 & 0.624 & 0.534     \\
\cline{2-8}
                      & \begin{tabular}{c} +\textbf{\textit{B}} \end{tabular}  & 0.334 & 0.334 & 0.363 & 0.430 & 0.587 & 0.495     \\
\cline{2-8}
                      & \begin{tabular}{c} +\textbf{\textit{B}},\textbf{\textit{C}} \end{tabular}  & 0.200 & 0.176 & 0.189 & 0.256 & 0.254 & 0.224     \\
\cline{2-8}
                      & \begin{tabular}{c} +\textbf{\textit{full}} \end{tabular}  & 0.186 & \textbf{0.164} & 0.176 & 0.241 & \textbf{0.244} & \textbf{0.209}     \\
\cline{2-8}
                      & \begin{tabular}{c} +\textbf{\textit{full}},\textbf{\textit{M}} \end{tabular}  & \textbf{0.185} & 0.165 & \textbf{0.173} & \textbf{0.240} & 0.247 & \textbf{0.209}     \\
\hline
\hline
\multirow{3}{*}{Traj} & \begin{tabular}{c} \textit{Sp} \end{tabular}  & 0.042 & 0.072 & 0.107 & 0.045 & 0.113 & 0.154     \\
\cline{2-8}
                      & \begin{tabular}{c} +\textbf{\textit{full}} \end{tabular}  & \textbf{0.039} & \textbf{0.069} & \textbf{0.096} & \textbf{0.043} & \textbf{0.108} & \textbf{0.147}     \\
\cline{2-8}
                      & \begin{tabular}{c} +\textbf{\textit{full}},\textbf{\textit{M}} \end{tabular}  & 0.040 & 0.071 & 0.098 & 0.044 & \textbf{0.108} & 0.151     \\
\cline{1-8}
\end{tabular}
}
\end{table}

\xwjrevise{\autoref{table:ablation_BBC-VINS} and \autoref{table:ablation_BBC-ORB} yield the following insights: 1) Integrating the depth predicted by BBC-Net into BBC-VINS optimization improves both depth accuracy and trajectory precision. 2) Base Balance Training and confidence enhance the accuracy of the network's predicted depth, with confidence showing a relatively more significant improvement. 3) Incorporating BBC-Net's predicted depth into BBC-ORBSLAM optimization primarily enhances depth accuracy. The notable numerical enhancement seen with Confidence, compared to Base Balance Training, is due to Confidence's ability to filter out areas with larger errors, such as in regions with weak textures. In contrast, base balance training primarily targets the removal of abnormal depth protrusions, as depicted in \autoref{fig:BBC-Net_improvement}. These anomalies usually represent a smaller portion and may not be as pronounced in the RMSE metric as Confidence. The reason why integrating the predicted depth into BBC-ORBSLAM does not enhance trajectory accuracy is that ORB-SLAM inherently maintains a comprehensive set of multi-frame visual associations, and the predicted depth struggles to provide additional multi-frame observational constraints to improve trajectory. 
Another pleasing discovery is that marginalization did not negatively impact performance. Therefore, we can confidently use marginalization to prevent rapid memory growth and accelerate optimization.}
\begin{table}
\centering
\caption{Evaluation of BBC-ORB(VI) on EuRoC. The best result is shown in bold. 
}
\label{table:ablation_BBC-ORB}
\scalebox{0.96}[1.0]{
\begin{tabular}{c|c|cccccc} 
\hline
    \multicolumn{2}{c}{} & V101 & V102 & V103 & V201 & V202 & V203 \\
\hline
\multirow{5}{*}{Depth} & \begin{tabular}{c} \textit{Sp} \end{tabular}  & 0.494 & 0.235 & 0.443 & 0.443 & 0.760 & 0.651     \\
\cline{2-8}
                      & \begin{tabular}{c} \textit{\textbf{\cite{qu2020depth}-R}}
                      \end{tabular}  & 0.379 & 0.264 & 0.268 & 0.396 & 0.582 & 0.338     \\
\cline{2-8}
                      & \begin{tabular}{c} +\textbf{\textit{B}} \end{tabular}  & 0.332 & 0.227 & 0.269 & 0.387 & 0.540 & 0.334     \\
\cline{2-8}
                      & \begin{tabular}{c} +\textbf{\textit{B}},\textbf{\textit{C}} \end{tabular}  & 0.215 & 0.155 & 0.190 & 0.245 & 0.329 & 0.206     \\
\cline{2-8}
                      & \begin{tabular}{c} +\textbf{\textit{full}} \end{tabular}  & \textbf{0.212} & \textbf{0.155} & \textbf{0.169} & \textbf{0.244} & \textbf{0.327} & \textbf{0.198}     \\
\hline
\hline
\multirow{2}{*}{Traj} & \begin{tabular}{c} \textit{Sp} \end{tabular}  & 0.033 & 0.011 & 0.013 & 0.018 & 0.013 & 0.019     \\
\cline{2-8}
                      & \begin{tabular}{c} +\textbf{\textit{full}} \end{tabular}  & 0.033 & 0.011 & 0.013 & 0.018 & 0.013 & 0.018     \\
\cline{1-8}
\end{tabular}
}
\end{table}

\xwjrevise{In addition to statistical analysis of numerical accuracy, we visualized mesh smoothness improvements resulting from the integration of each module in \autoref{fig:ablation_study_v12} and \autoref{fig:ablation_study_v23_orbvi}. Enhanced mesh smoothness signifies improved trajectory or depth consistency. These visual improvements align with the numerical analysis presented in \autoref{table:ablation_BBC-VINS} and \autoref{table:ablation_BBC-ORB}.
From \autoref{fig:ablation_study_v12} and \autoref{fig:ablation_study_v23_orbvi}, it is evident that directly incorporating~\cite{qu2020depth} into the SLAM process generates a mesh with significant noise. With the refinement of depth-based balancing, the mesh becomes smoother. Confidence measures contribute to a substantial reduction in large noisy areas. Furthermore, after BA optimization, the issue of uneven mesh is almost entirely resolved.}

\subsection{Online Demo on a Mobile Phone}
To \xwjrevise{demonstrate the effectiveness} of our proposed method, we have \xwjrevise{implemented a mobile application}. This application takes an RGB image stream and IMU measurements from the device as input\xwjrevise{, utilizing BBC-VINS to estimate poses and depths}. The predicted depth of each keyframe is then fused into a global TSDF model using the tracked 6DoF pose in a separate fusion thread. Simultaneously, the fused TSDF volume is raycasted by the viewing pose on the frontend, providing users with a real-time preview of the reconstructed mesh. For more details, please refer to the supplementary video.
\xwjrevise{Our fusion method relies on Mobile3DRecon~\cite{yang2020mobile3drecon}. Additionally, we adopt the strategy outlined in~\cite{xiang2021mobile3dscanner} to prevent excessive memory growth. The depths optimized through BA are subsequently reintegrated into the TSDF model.}

\subsection{Runtime Measurement}
\label{sec:runtime_analyze}

\setlength{\tabcolsep}{0.8mm}
\begin{table}
\centering
\caption{Evaluation of average time(ms) per frame for each module. \xwjrevise{Results with * are tested on mobile, while others are tested on PC.}}
\label{table:vs_efficiency}
\scalebox{0.95}[1.0]{
\begin{tabular}{c|ccccccc} 
\hline
            & \begin{tabular}[c]{@{}c@{}}Ours*\end{tabular}
            & \begin{tabular}[c]{@{}c@{}}NICE\cite{zhu2022nice}\end{tabular}
            & \begin{tabular}[c]{@{}c@{}}Tandem\end{tabular} 
            & \begin{tabular}[c]{@{}c@{}}CF\cite{zhang2023completionformer}\end{tabular}
            & \begin{tabular}[c]{@{}c@{}}CodeVIO\end{tabular}
            & \begin{tabular}[c]{@{}c@{}}CodeMaping\end{tabular} 
            \\ 
\hline
\xwjminor{Param} 
& \xwjminor{28.59M} 
& \xwjminor{11.92M} & \xwjminor{0.89M} & \xwjminor{79.63M} & \xwjminor{-} & \xwjminor{-}            
\\ 
\hline
Infer(per iter) 
& 20 
& 209 & 513 & 265 & - & 11            
\\ 
\hline
Tracking 
& 16.9 
& 3118 & 91 & - & 44 & -             
\\ 
\hline
Optimize 
& 84 
& 3458 & 16 & - & - & 170
\\
\hline
\end{tabular}
}
\end{table}
\xwjrevise{
Due to the unavailability of source code for several methods(CodeVIO and CodeMapping) and deployment constraints, the values in \autoref{table:vs_efficiency} are not obtained on the same device. Our method(BBC-VINS)'s timing is based on iPhone 12 Pro, while timings for other methods were recorded on a PC (NICE-SLAM, CompletionFromer, CodeVIO on 1080Ti GPU, Tandem on 1060 GPU, CodeMapping on 3080 GPU). CodeMapping employs a window size of 4 for window optimization, whereas our method utilizes a window size of 8. Generally, depth completion networks tend to achieve faster speeds compared to multi-frame depth prediction networks like MVSNet (Tandem). CodeMapping showing slightly faster than ours, mainly due to device differences. In terms of network structure alone, our network's computational complexity should be comparable to that of CodeMapping. Given that BBC-Net does not incorporate an SPN layer, it is reasonable to expect our network to have a noticeable advantage in speed over networks with SPN, such as CompletionFormer. 
The reason why Tandem optimization takes less time is that it does not optimize depth.
Indeed, for systems where map optimization and network inference are parallelized, such as our method, CodeMapping, and Tandem, the efficiency of dense mapping is determined by the bottleneck of optimization speed and network inference speed. Therefore, even when comparing our method's evaluation time on a mobile with the reported times of other methods on a PC, our method still demonstrates significant speed advantages in terms of efficiency.
}

 \section{Conclusion}

In this paper, we propose BBC-Net, an effective depth completion network specifically designed for coupling with the SLAM framework. By utilizing the predicted multi-basis depths and confidence, we incorporate the weights of bases into the optimization of the SLAM framework. We have designed multiple factors to optimize the weight of depth bases, which can be flexibly combined based on the actual scenario. Additionally, our method demonstrates high efficiency, making it particularly suitable for real-time dense SLAM. We have also successfully adapted our method for mobile devices and verified its performance. \xwjrevise{The experiment shows that our approach outperforms the state-of-the-art both in accuracy and efficiency.}

\section*{Acknowledgments}
This work was partially supported by NSF of China (No. 61932003).


\begin{IEEEbiography}[{\includegraphics[width=1in,height=1.25in,clip,keepaspectratio]{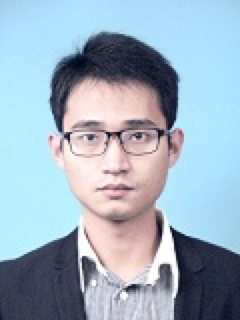}}]{Weijian Xie}
 is currently a D.Eng. student of the State Key Lab of CAD\&CG, Zhejiang University, advised by Prof. Hujun Bao and Prof. Guofeng Zhang. He received his Master's degree in computer science from Zhejiang University in 2017. He is also working as Senior Algorithm Scientist at SenseTime Research. His research interests include SLAM, 3D reconstruction, and augmented reality.\end{IEEEbiography}

\begin{IEEEbiography}[{\includegraphics[width=1in,height=1.25in,clip,keepaspectratio]{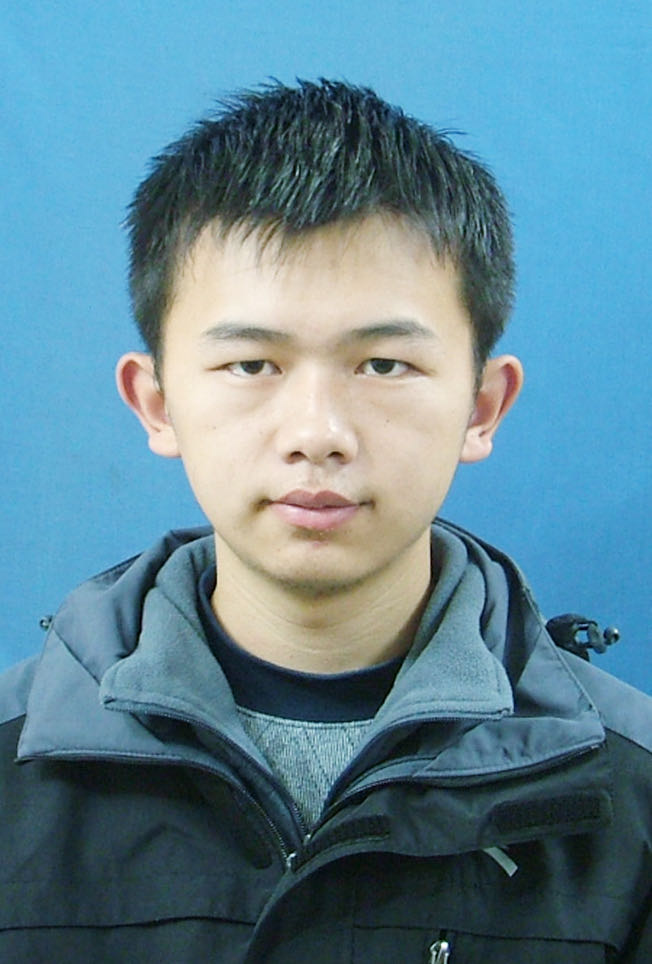}}]{Guanyi Chu}
 is currently working as a Computer Visual Researcher at SenseTime Research. He obtained his Master's degree in the State Key Lab of CAD\&CG, Zhejiang University in 2019. His research interests include SLAM and the application of deep learning in 3D vision.\end{IEEEbiography}

\begin{IEEEbiography}[{\includegraphics[width=1in,height=1.25in,clip,keepaspectratio]{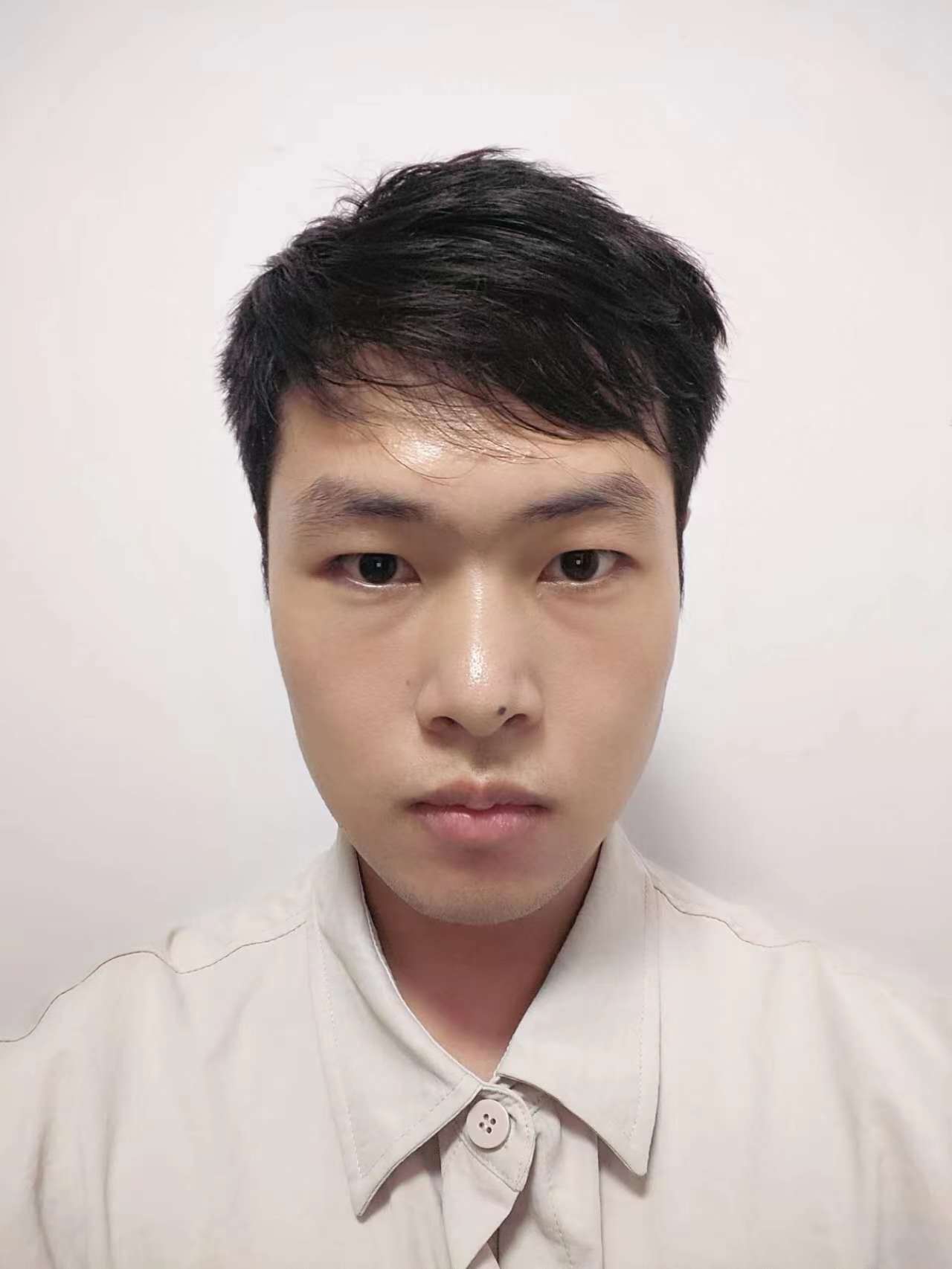}}]{Quanhao Qian}
is currently working as a Computer Visual Researcher at SenseTime Research. Prior to this, he obtained his master's degree in the CAD \& CG National Key Laboratory at Zhejiang University in 2019. His primary research interests lie in the fields of SLAM and the application of deep learning in 3D vision.\end{IEEEbiography}

\begin{IEEEbiography}[{\includegraphics[width=1in,height=1.25in,clip,keepaspectratio]{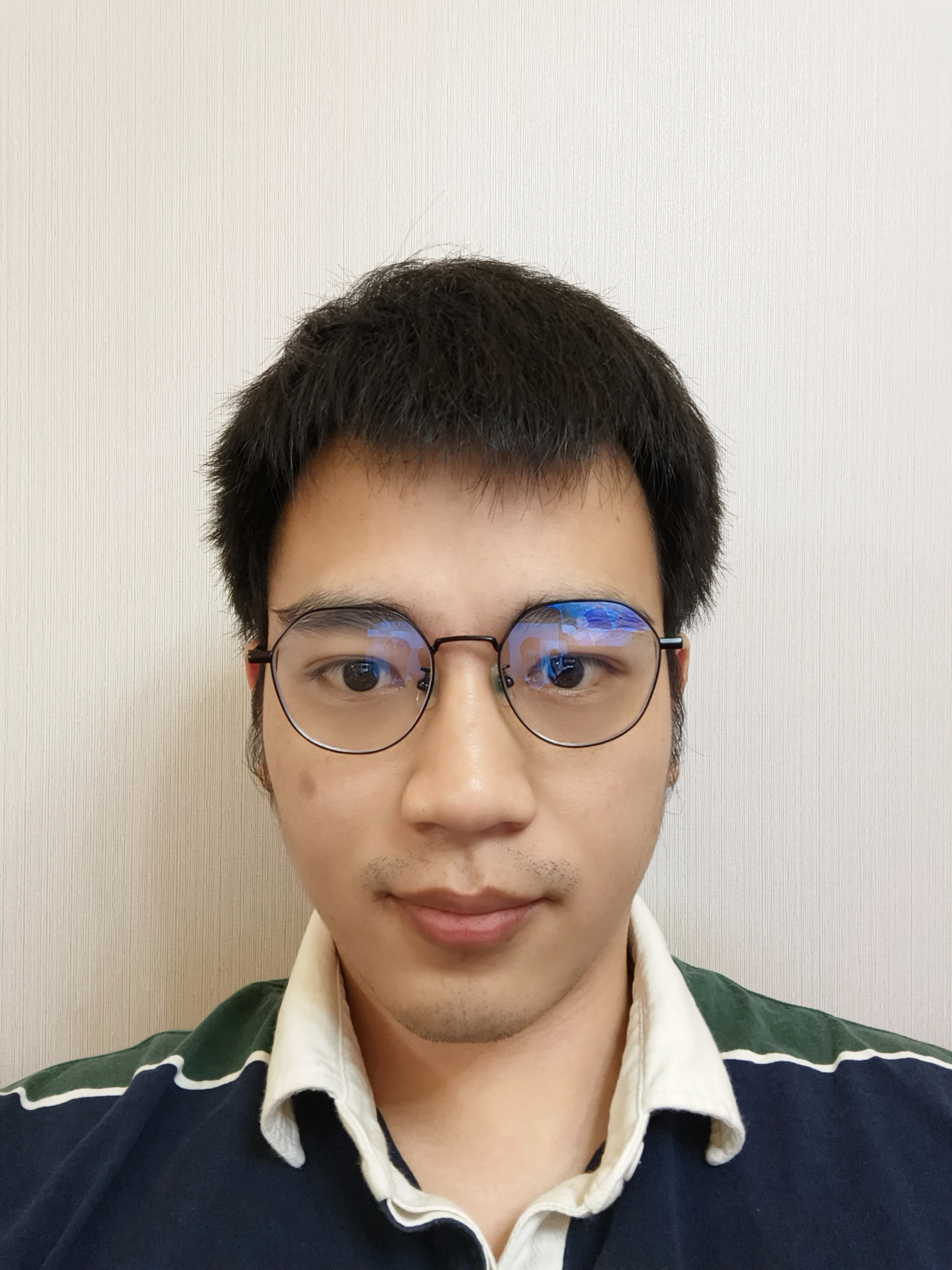}}]{Yihao Yu}
is currently working as a Computer Visual Researcher at SenseTime Research. He obtained his Master's degree in Computer science and Technology from Beihang University in 2019. His research interests include 3D reconstruction and pose estimation.\end{IEEEbiography}

\begin{IEEEbiography}[{\includegraphics[width=1in,height=1.25in,clip,keepaspectratio]{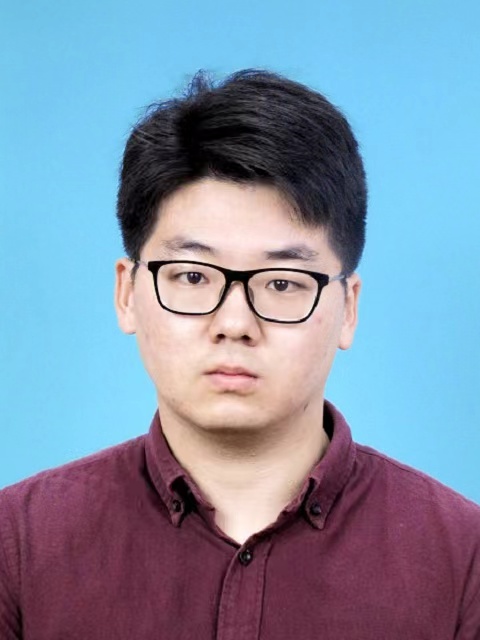}}]{Hai Li}
 received the B.S. degree in Computer Science and Technology from Harbin Engineering University in 2016. Subsequently, he obtained Ph.D. degree from Zhejiang University in 2023. His research interests include 3D reconstruction, SLAM and Augmented Reality.\end{IEEEbiography}

\begin{IEEEbiography}[{\includegraphics[width=1in,height=1.25in,clip,keepaspectratio]{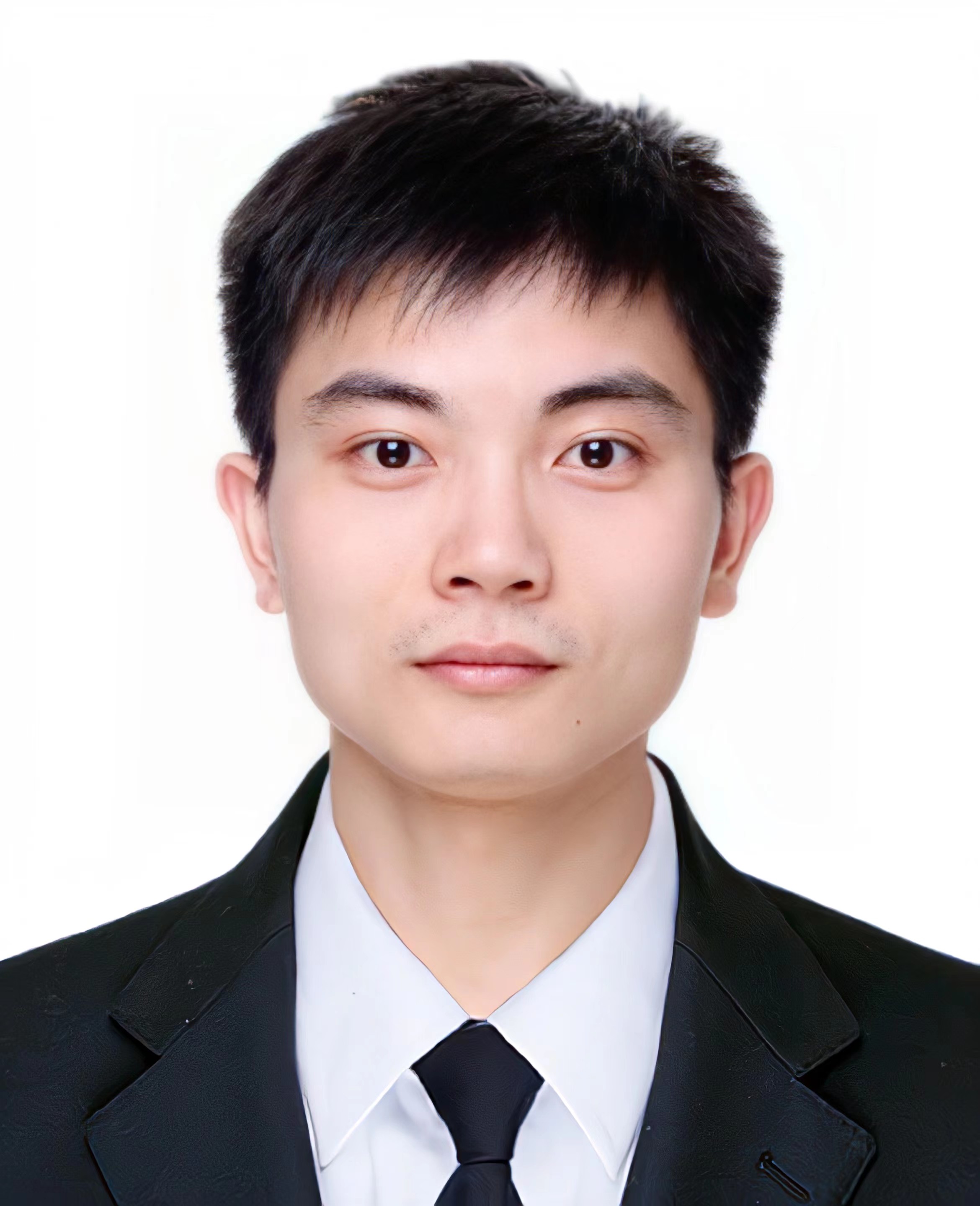}}]{Danpeng Chen}
is a D.Eng. student of the State Key Lab of CAD\&CG, Zhejiang University, advised by Prof. Hujun Bao and Prof. Guofeng Zhang. He also worked at SenseTime \& Tetras.ai. His research interests include SLAM, 3D reconstruction, and their applications in virtual augmented reality.\end{IEEEbiography}

\begin{IEEEbiography}[{\includegraphics[width=1in,height=1.25in,clip,keepaspectratio]{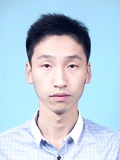}}]{Shangjin Zhai}
is currently working as a Computer Visual Researcher at SenseTime Research. He obtained his Master's degree in the State Key Lab of CAD\&CG, Zhejiang University, where he specialized in the research area of Simultaneous Localization and Mapping (SLAM).\end{IEEEbiography}

\begin{IEEEbiography}[{\includegraphics[width=1in,height=1.25in,clip,keepaspectratio]{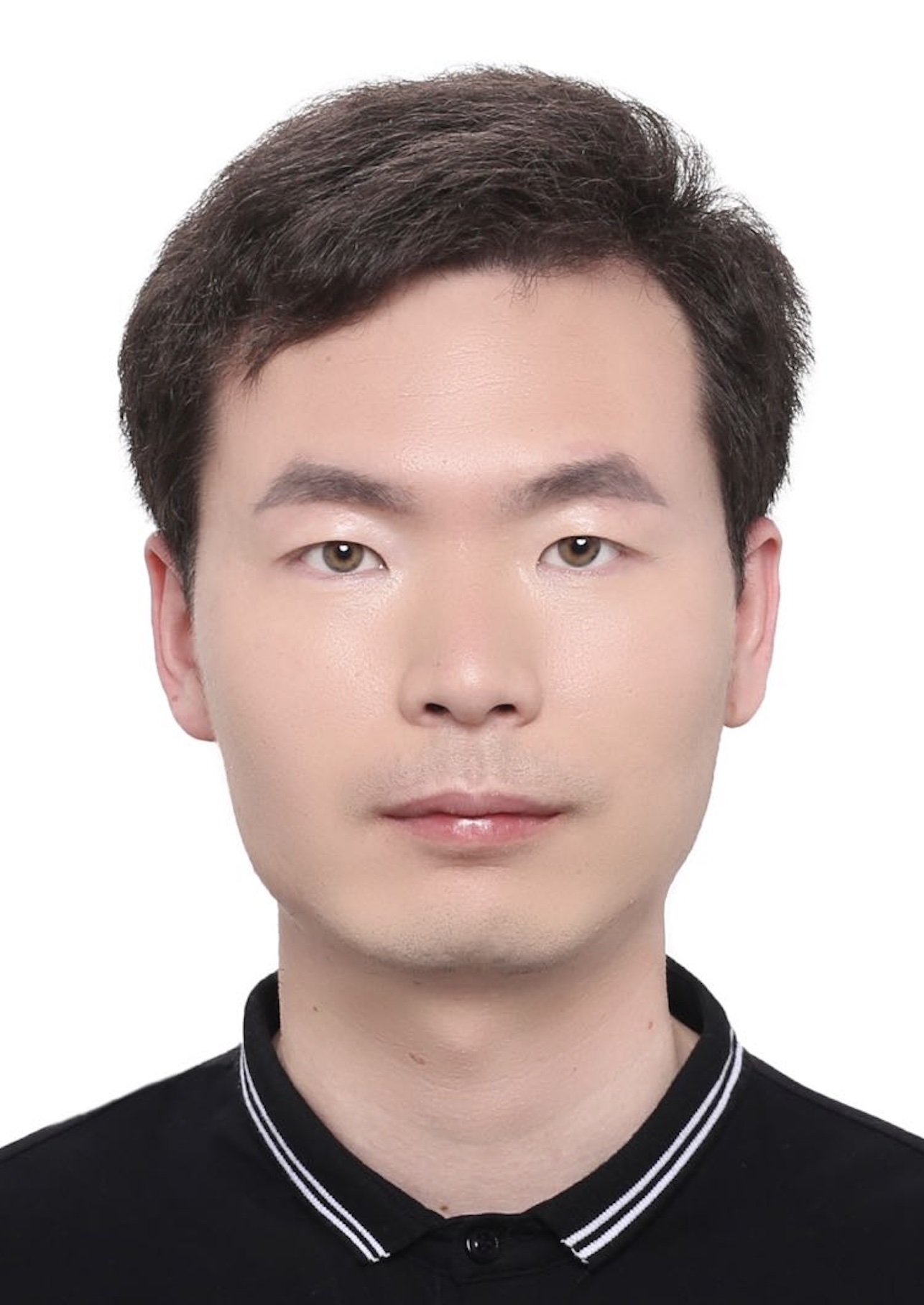}}]{Nan Wang}
is currently working as Associate Research Director at SenseTime \& Tetras.ai. Before that, he was a Senior R\&D at Baidu Inc.. He obtained his Master's degree in the State Key Lab of CAD\&CG, Zhejiang University in 2015, advised by Prof. Guofeng Zhang. His research interests include SLAM, 3D reconstruction, and augmented reality.\end{IEEEbiography}

\begin{IEEEbiography}[{\includegraphics[width=1in,height=1.25in,clip,keepaspectratio]{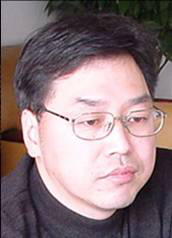}}]{Hujun Bao}
 (Member, IEEE) is currently a Professor at the Computer Science Department, Zhejiang University, and the former Director of the State Key Laboratory of Computer Aided Design and Computer Graphics. He leads the Mixed Reality Group, State Key Laboratory of Computer Aided Design and Computer Graphics, to make a wide range of research on 3D reconstruction and modeling, real-time rendering and virtual reality, real-time 3D fusion, and augmented reality. Some of these algorithms have been successfully integrated into the mixed reality system SenseMARS. His research interests include computer graphics, computer vision, and mixed reality.\end{IEEEbiography}

\begin{IEEEbiography}[{\includegraphics[width=1in,height=1.25in,clip,keepaspectratio]{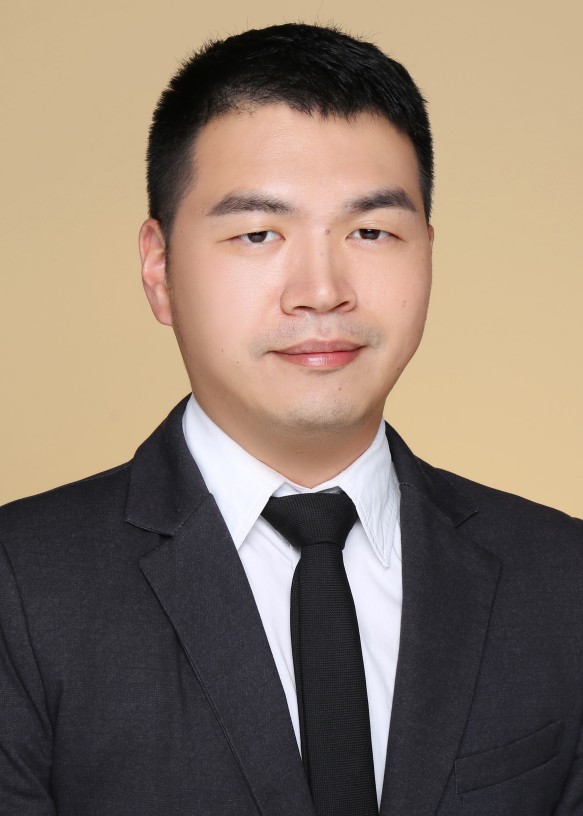}}]{Guofeng Zhang}
(Member, IEEE) received the B.S. and Ph.D. degrees in computer science and technology from Zhejiang University in 2003 and 2009, respectively. He is currently a Professor at Zhejiang University. His research interests include SLAM, 3D reconstruction, and augmented reality. He received the National Excellent Doctoral Dissertation Award, the Excellent Doctoral Dissertation Award of the China Computer Federation, and the ISMAR 2020 Best Paper Award.\end{IEEEbiography}

\end{document}